%% file: main.tex
\documentclass[sigconf]{acmart}
\AtBeginDocument{%
  \providecommand\BibTeX{{%
    \normalfont B\kern-0.5em{\scshape i\kern-0.25em b}\kern-0.8em\TeX}}}

\setcopyright{acmcopyright}
\copyrightyear{2018}
\acmYear{2018}
\acmDOI{XXXXXXX.XXXXXXX}

\acmConference[Conference acronym 'XX]{Make sure to enter the correct
  conference title from your rights confirmation emai}{June 03--05,
  2018}{Woodstock, NY}
\acmPrice{15.00}
\acmISBN{978-1-4503-XXXX-X/18/06}

\usepackage{algorithm2e}
\SetKwInput{KwInput}{Input}
\SetKwInput{KwOutput}{Output}
\SetKwInput{KwReturn}{Return}
\usepackage{subfig}
\input{math_commands.tex}
\begin{document}

%%
%% The "title" command has an optional parameter,
%% allowing the author to define a "short title" to be used in page headers.
\title{Self-Supervised Contrastive Pre-Training for Multivariate Point Processes}

%%
%% The "author" command and its associated commands are used to define
%% the authors and their affiliations.
%% Of note is the shared affiliation of the first two authors, and the
%% "authornote" and "authornotemark" commands
%% used to denote shared contribution to the research.
\author{Xiao Shou}
% \authornote{Both authors contributed equally to this research.}
% \orcid{1234-5678-9012}
% \author{G.K.M. Tobin}
% % \authornotemark[1]
% \email{webmaster@marysville-ohio.com}
\affiliation{%
  \institution{RPI}
  % \streetaddress{P.O. Box 1212}
  \city{Troy}
  \state{NY}
  \country{USA}
  % \postcode{43017-6221}
}
% \email{xshou01@gmail.com}

\author{Dharmashankar Subramanian}
\affiliation{%
  \institution{IBM AI Research}
  % \streetaddress{1 Th{\o}rv{\"a}ld Circle}
  \city{Yorktown Heights}
  \state{NY}
  \country{USA}}
% \email{larst@affiliation.org}

\author{Debarun Bhattacharjya}
\affiliation{%
  \institution{IBM AI Research}
  % \streetaddress{1 Th{\o}rv{\"a}ld Circle}
  \city{Yorktown Heights}
  \state{NY}
  \country{USA}}

\author{Tian Gao}
\affiliation{%
  \institution{IBM AI Research}
  % \streetaddress{1 Th{\o}rv{\"a}ld Circle}
  \city{Yorktown Heights}
  \state{NY}
  \country{USA}}  

\author{Kristin P. Bennett}
% \authornote{Both authors contributed equally to this research.}
% \orcid{1234-5678-9012}
% \author{G.K.M. Tobin}
% % \authornotemark[1]
% \email{webmaster@marysville-ohio.com}
\affiliation{%
  \institution{RPI}
  % \streetaddress{P.O. Box 1212}
  \city{Troy}
  \state{NY}
  \country{USA}
  % \postcode{43017-6221}
}

%%
%% By default, the full list of authors will be used in the page
%% headers. Often, this list is too long, and will overlap
%% other information printed in the page headers. This command allows
%% the author to define a more concise list
%% of authors' names for this purpose.
\renewcommand{\shortauthors}{Shou et al.}

%%
%% The abstract is a short summary of the work to be presented in the
%% article.
\begin{abstract}
Self-supervision is one of the hallmarks of representation learning in the increasingly popular suite of foundation models including large language models such as BERT and GPT-3, but it has not been pursued in the context of multivariate event streams, %i.e. data with irregular occurrences of different event types,
%datasets involving irregular occurrences of different event types,
to the best of our knowledge.
We introduce a new paradigm for self-supervised learning for %temporal 
\textcolor{black}{multivariate} point processes %, i.e. data with irregular occurrences of different event types,
using a transformer encoder. Specifically,
we design a novel pre-training strategy for the encoder % called masked event modeling
where we not only mask random event epochs but also insert randomly sampled `void' epochs where an event does not occur; this differs from the typical discrete-time pretext tasks such as word-masking in BERT but expands the effectiveness of masking to better capture continuous-time dynamics. To improve downstream tasks, we introduce a contrasting module that compares real events to simulated void instances. %To learned better presentation for downstream tasks, we introduce a new contrasting module built upon the real events and simulated void instances.  
%but prevents the immediate adoption of discrete-time pretext algorithms.
% on a potentially larger dataset and then fine-tuned on 
%A unique aspect of event sequence - void time where no events happen between two consecutive epochs is governed by continuous time dynamics and prevent the immediate adaption of discrete time pretext algorithms. To this end, we design a novel pretraining strategy called masked event modeling where we not only mask random event epochs but also insert randomly sampled void epochs for masking. 
The pre-trained model can subsequently be fine-tuned on a potentially much smaller event dataset, similar conceptually to the typical transfer of popular pre-trained language models.
We demonstrate the effectiveness of our proposed paradigm on the
next-event prediction task using synthetic datasets and 3 real applications, observing a relative performance boost of as high as up to 20\% compared to state-of-the-art models.  
\end{abstract}

% \ccsdesc[500]{Do Not Use This Code~Generate the Correct Terms for Your Paper}
% \ccsdesc[300]{Do Not Use This Code~Generate the Correct Terms for Your Paper}
% \ccsdesc{Do Not Use This Code~Generate the Correct Terms for Your Paper}
% \ccsdesc[100]{Do Not Use This Code~Generate the Correct Terms for Your Paper}

%%
%% Keywords. The author(s) should pick words that accurately describe
%% the work being presented. Separate the keywords with commas.

%% A "teaser" image appears between the author and affiliation
%% information and the body of the document, and typically spans the
%% page.
% \begin{teaserfigure}
%   \includegraphics[width=\textwidth]{sampleteaser}
%   \caption{Seattle Mariners at Spring Training, 2010.}
%   \Description{Enjoying the baseball game from the third-base
%   seats. Ichiro Suzuki preparing to bat.}
%   \label{fig:teaser}
% \end{teaserfigure}

\received{20 February 2007}
\received[revised]{12 March 2009}
\received[accepted]{5 June 2009}

%%
%% This command processes the author and affiliation and title
%% information and builds the first part of the formatted document.
\maketitle

\section{Introduction}

% Para on transfer, self-supervision and foundation models
Transfer learning occurs when a model is pre-trained on a task, such as classification on a large labelled dataset such as ImageNet, and the model's `knowledge' is then applied to another task, such as classification on medical images. In the current era of AI, transfer in domains such as natural language processing and image processing  often leverages \emph{self-supervised learning}, where pre-training for representation learning is done using unlabeled data. Although the fundamental ideas of transfer are not new, there is a clear emerging paradigm around foundation models~\citep{bommasani}, such as BERT \citep{devlin2018bert} and GTP-3 \citep{brown2020language}, which are trained with diverse unlabeled data at scale using self-supervision. These pre-trained models are then fine-tuned and adapted to different downstream tasks that respectively come with limited labeled data. Recent progress has been possible primarily due to improvements in  hardware, development of the attention mechanism~\citep{vaswani2017attention}, and availability of substantial unlabeled training data.

\begin{figure} [htbp]
    \centering
{{\includegraphics[width=6.cm]{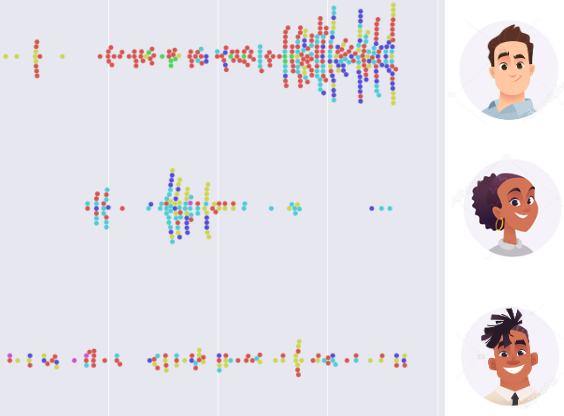} }}%
    \caption{An example of real-world decentralized finance transactions from 3 users where their time-stamped actions are events displayed using colored markers. The 6 types of events are: borrow (red), repay (yellow), liquidation (green), deposit (blue), redeem (purple) and swap (pink). %Their timestamped actions are captured and displayed. 
    }%
    \label{fig:example}%
\end{figure}

% Event models
We extend and pursue self-supervised learning in the context of \emph{multivariate event streams}, i.e. data involving irregular occurrences of different types of events. Event stream datasets are widely available across domains, for instance in the form of social network interactions, customer transactions, system logs, financial events, health episodes, etc. An example of decentralized finance transactions from 3 users is shown in Figure \ref{fig:example}. It is well known that \emph{temporal point processes} provide a sound mathematical framework for modeling such datasets \citep{daley2003introduction}. An interesting application of temporal point processes is well-timed recommendation systems \cite{kim2017read}. In this paper, we introduce a new paradigm for
self-supervised learning for temporal point processes using a transformer encoder. 
Although self-supervised learning has recently been explored for classical time series data~\textcolor{black}{\citep{zerveas2021transformer,zhang2022self}},
to the best of our knowledge, self-supervision has not yet been explored in the context of point processes. 

% Distinguish from prior work
Neural models for temporal point processes~\citep[e.g.][]{du2016recurrent,mei2016neural,xiao2017modeling} have advanced the state of the art in event modeling, particularly for the task of event prediction. The typical approach in this line of work is to train a neural network on a large amount of event data. 
Our proposed paradigm differs from current standard practices % of training (neural) MTPP models, namely log-likelihood maximization via neural conditional intensity functions, which rely solely on supervised training from a single large event dataset. Instead, 
by taking a \textcolor{black}{\textit{transfer learning}} approach analogous to foundation models: to first pre-train a neural model on a (potentially) large event dataset and then fine-tune the model for prediction on a limited event dataset. 
%We consider the typical setting of homogeneous transfer learning~\citep{9134370}, where all datasets that get pooled for pre-training involve  the same set of event types. This allows for potentially leveraging different datasets even though there may be realistic variations with respect to parametric or structural dependencies. %present in the corresponding event streams in each data set.
%For e.g., consider several data sets, each containing multiple event streams over a common label set, say $\mathcal{L} = \{a,b,c\}$, where each data set corresponds to a distinct multivariate Hawkes process with its own infectivity matrix and excitation parameters. 

%We leave the more complex case of heterogeneous transfer learning with data sets that contain non-overlapping labels or event types to future work. [COMMENT from DB: I have moved this statement to the concluding section.]

% Motivate transfer learning applications and mention multi-task literature
Transfer learning with event models has many potential applications. For instance, there may be abundant data from electronic health records containing information around a particular patient population; this data could potentially be leveraged for another population whose data is either unavailable or harder to obtain. This is an issue relevant to health equity since there may be data related concerns  for some under represented populations. 
Similarly, financial event data from an industrial sector could potentially be transferred to another.
Electronic commerce is yet another illustrative domain where transfer learning techniques may help to transfer purchase behaviors across a large pool of user populations and/or product types.

%For example,  it can find useful applications for analyzing electronic health records that contain patient data for different patients and diseases. Tasks of interest include, given a patient's history of visits up until time $t$, a) when will the next patient visit happen, and b) how many visits are expected over an interval $(t, t+T]$. Since typical  data of a single patient or a given disease type tends to be rare and sparse, transfer learning techniques are pertinent to address these questions in order to improve resource allocation strategies in health care \citep{amarasingham2010automated}. Likewise, in electronic commerce, transfer learning techniques may help to learn user purchase behaviors across a large pool of individual users, each with limited click and purchase data, for improved recommendations.

%  Prior work + position our work
Although there is some related work around multi-task learning with event streams, such as through deploying hierarchical Gaussian process models~\citep{lian2015multitask} %that proposed a to transfer information across multiple tasks. In some recent work, a
or time-scale graphical event models~\citep{monvoisin2019multi}, this line of research typically considers learning by pooling together disparate data from the same population. % for instance, transferring 
%Furthermore, efficient inference remains a computational challenge in such graphical models.
In contrast, we tackle the more ambitious effort of transferring from one or multiple event datasets to another. 
Specifically, we consider the typical setting of homogeneous transfer learning~\citep{9134370}, where all datasets that get pooled for pre-training involve the same set of event types. This type of transfer is not overly simplified; in fact, much of transfer learning research focuses on this type of transfer. This setting allows for potentially leveraging different datasets even though there may be realistic variations with respect to parametric or structural dependencies present in the corresponding event streams across each dataset. In our multivariate event streams case,  the datasets for pre-training and transfer may involve very different underlying dynamics.
All the examples mentioned in above paragraph are applications that could benefit from homogeneous transfer.

%For e.g., consider several data sets, each containing multiple event streams over a common label set, say $\mathcal{L} = \{a,b,c\}$, where each data set corresponds to a distinct multivariate Hawkes process with its own infectivity matrix and excitation parameters. 

%In contrast, we propose techniques for achieving effective self-supervision in point process data via a novel pretext task, and present the first such work that is analogous to foundation models for point process data sets.

%[Should include time series paper somewhere in related work]

%\subsection{leverage domain information and background knowledge}

%We can generate synthetic data; additional domain knowledge maybe insightful, however as the name suggest, it requires expert carefully examine cases. For example temporal logic rules for when disease A occurs before B, such rules are carefully constructed or validated by domain experts \citep{li2020temporal,li2021explaining}. Other statements such as event A depends on event B in graphical event models also may not be directly accessible to an analyst. 

\textbf{Contributions}: We make the following major contributions: % for self-supervised learning for event datasets. 
% \begin{itemize}
1) We introduce a self-supervised paradigm for transfer learning in temporal point processes; a crucial innovation is to explicitly incorporate information about the absence of events, which improves the modeling of temporal dynamics without burdening training efficiency. 
%with void events. This simple strategy improves the modeling of temporal dynamics without burdening the training efficiency.
2) We propose a masked event model,  which is a new way to derive a pretext task for self-supervision targets in transformer models for event streams. 
%To effectively use void events, we propose a masked event model, which is a new way to derive a pretext task for self-supervision targets in transformer models. 
3) We introduce a novel contrasting module that leverages the (dis)similarity between real instances and simulated void events to improve downstream prediction tasks.
4) We conduct an empirical evaluation that demonstrates improved transfer learning performance for event prediction on synthetic and real datasets, relative to state-of-the-art transformer event models \textcolor{black} {and those leveraging noise contrastive estimation}. 
% \end{itemize}

\section{Background and Related Work}
\label{bkg}

%\subsection{MTPPs and NTPPs}

\subsection{Temporal Point Processes}

Multivariate temporal point processes (MTPP) are elegant mathematical models for event streams where  event types/labels from some discrete set occur in continuous time \citep{daley2003introduction}. A multi-dimensional MTPP generates sequences with time stamps and associated labels of the form $S = \{(t_i , y_i)\}_{i=1}^n$ where $t_i$ is the time of occurrence of $i^{th}$ event and $y_i$ is its label. The cardinality of label set $\mathbb{L}$ is $M$. A strict temporally ordered event stream assumes a period of events observed within $[0,T]$ for each  $t_i \in [0,T] $ for all $i \in [1,2,...,n]$. %Generally, a univarite TPP can be characterized by its conditional intensity function (CIF) which intuitively depicts the rate of events occurring at a specific time $t$, 
MTPPs are characterized by conditional intensity functions for each label representing the rate at which it occurs at any time $t$, $\lambda_e (t) = \lim_{\Delta t \rightarrow 0} \frac{\E(\mathcal{N}_e(t+\Delta t |h_t) - \mathcal{N}_e(t |h_t) )}{\Delta t}$,
%\begin{equation}
%    \lambda (t) = \lim_{\Delta t \rightarrow 0} \frac{\E(\mathcal{N}(t+\Delta t |h_t) - \mathcal{N}(t |h_t) )}{\Delta t}
%\end{equation}
where $\mathcal{N}_e(t |h_t)$ counts the number of occurrences of label $e$ prior to historical occurrences (or simply history) $h_t$. %MTPP on the other hand models CIF for each event type $\lambda_e (t)$. 

Several MTPP models such as multivariate versions of the classic Hawkes process \citep{hawkes1971spectra} and piecewise-constant models~\citep{gunawardana2016universal,bhattacharjya2018proximal} assume some parametric form of the conditional intensity function.  Neural MTPPs are more recent variants that capture the underlying dynamics using neural networks. Recent years have witnessed rising popularity of neural MTPPs due to their state-of-the-art performance on benchmark datasets for predictive tasks~\citep{du2016recurrent,mei2016neural,xiao2017modeling, omi2019fully,shchur2019intensity,zuo2020transformer}. A common training objective in neural MTPPs involves minimizing the negative log-likelihood. The log-likelihood of observing a sequence $S$ is the sum of log-likelihood of events and non-events and can be computed as:
\begin{equation} \label{eqn:ll}
\text{log} ~ p(S) =  \sum_{i=1}^{n} \sum_{e=1}^{M} \mathbb{I}[y_i =e] \text{log} \lambda_e(t_i) - \int_{0}^T \sum_{e=1} ^M \lambda_e(t)dt
\end{equation}

%Neural Models: RMTPP, ERPP, NHP, Fully, LNM,

Many of these neural MTPPs assume some form of evolution dynamics between events in order to compute the second term in Eq.~\ref{eqn:ll}, such as recurrent neural network (RNN) evolution~\citep{xiao2017modeling}, exponential decay~\citep{mei2016neural}, intensity-free modeling of the integral~\citep{omi2019fully}, or the usage of explicit epochs indicating absence of events~\citep{gao2020multi}.

\subsection{Transformers for Event Data}

Attention~\citep{xiao2019learning} and transformer-based event models have shown promising results in recent years, including the self-attentive Hawkes process~\citep{zhang2020self}, transformer Hawkes process (THP)~\citep{zuo2020transformer}, attentive neural point process~\citep{gu2021attentive} and attentive neural datalog through time~\citep{mei2022transformer}. 
%[COMMENT: Xiao, whats these two? (A)-NHP, (A)-NDTT, xiao: these are the jason eisner paper iclr 22, nhp is neural hawkes, they implemented attention version of NHP, NDTT is the database paper by same author, icml 20 paper. ]
%Attention and transformer models have been used to model event data in recent years \citep{xiao2019learning,zhang2020self,zuo2020transformer,gu2021attentive}.
The self-attention mechanism, in our context, relates different event instances of a single stream in order to compute a representation of the stream. The architecture of transformers for MTPPs generally consists of an embedding layer and a self-attention layer. In THP, for example, the embedding layer embeds an event instance as a combination of temporal embedding \citep{zuo2020transformer} different from  position embedding \citep{vaswani2017attention} for language models, and one-hot encoded event-type; such representations of event sequences are then fed into attention blocks (each consists of dot-product attention and point-wise feed forward neural network) to output high-level representations (typically denoted as $\mathbf{H}_i$'s) for modeling conditional intensity functions.   

\subsection{Self-supervision for Sequence Data}

%[This is to explain prior work on sequences, i.e. discrete-time data. This includes language models like BERT but also new work in time series]

Sequence models such as RNNs have achieved much success in various applications, but more recent methods typically rely on transformer architectures \citep{vaswani2017attention} and the attention mechanism, specially in natural language processing. After  fine-tuning on downstream tasks, these pre-trained models %, which include BERT \citep{devlin2018bert}, GPT-2 \citep{radford2019language}, and many more, 
lead to sizable improvement over previous state of the art. However, large-scale transformers are also bulky and resource-hungry, typically with billions of parameters \citep{brown2020language} and  cost millions of USD to train \citep{floridi2020gpt}. 

Self-supervision is typically achieved in sequence models by deriving an effective pretext task that is trained through supervised learning with a masking strategy for indicating self-supervision targets. %\textcolor{blue}{In the self-supervision training stage, a masking strategy is generally used to indicate self-supervision targets}. 
For example, in BERT, about 15\% of the words are randomly masked using an independent Bernoulli model of masking, and replaced with a new \textit{[MASK]} label or a random word. In some recent work on self-supervision for time series data~\citep{zerveas2021transformer}, the masking is done to ensure longer lengths of masked values, all replaced with the value 0, to get geometrically distributed run-lengths of masked values. \textcolor{black}{A more recent self-supervised contrastive pre-training scheme for time series \cite{zhang2022self} leverages the time-frequency consistency of time series data and 
embeds a time-based neighborhood of an example close to its frequency-based neighborhood. To the best of our knowledge, no self-supervised contrastive pre-training has been proposed for multivariate point processes.}
Here we introduce a novel pretext task with a masking strategy specially tailored for asynchronous event data in continuous time. As we show later through experimental ablation studies, straightforward application of prior discrete-time sequence-based masking strategies proves inadequate for event data, because the intensity rates of events that represent continuous-time dynamics may vary in general between any two consecutively observed events. This aspect distinguishes event streams from discrete-time data such as time series and has not been addressed by masking in standard temporal transformers. %We propose to use a non-parametric approach of using void events to address the dynamic evolution of event intensities in transformers and then develop a corresponding masking  strategy for training.
% [COMMENT by DB: void events have not been defined yet!]

% \subsection{Contrastive Learning for Temporal Point Process}
% \textcolor{red}{while many constrastive learning framework for time series, most recent one "Self-Supervised Contrastive Pre-Training For Time Series via Time-Frequency Consistency" published in NeurIPS 2022.
% mention Initiator (Guo et al.ijcai 2018), NCE-TPP (mei et al neurips 2020), Hierarchical Contrastive Learning for Temporal Point Processes (Wang et al AAAI 23) }

\section{A Self-supervised Learning Paradigm}

\begin{figure*} [t]
    \centering
    \includegraphics[width=0.79\textwidth]{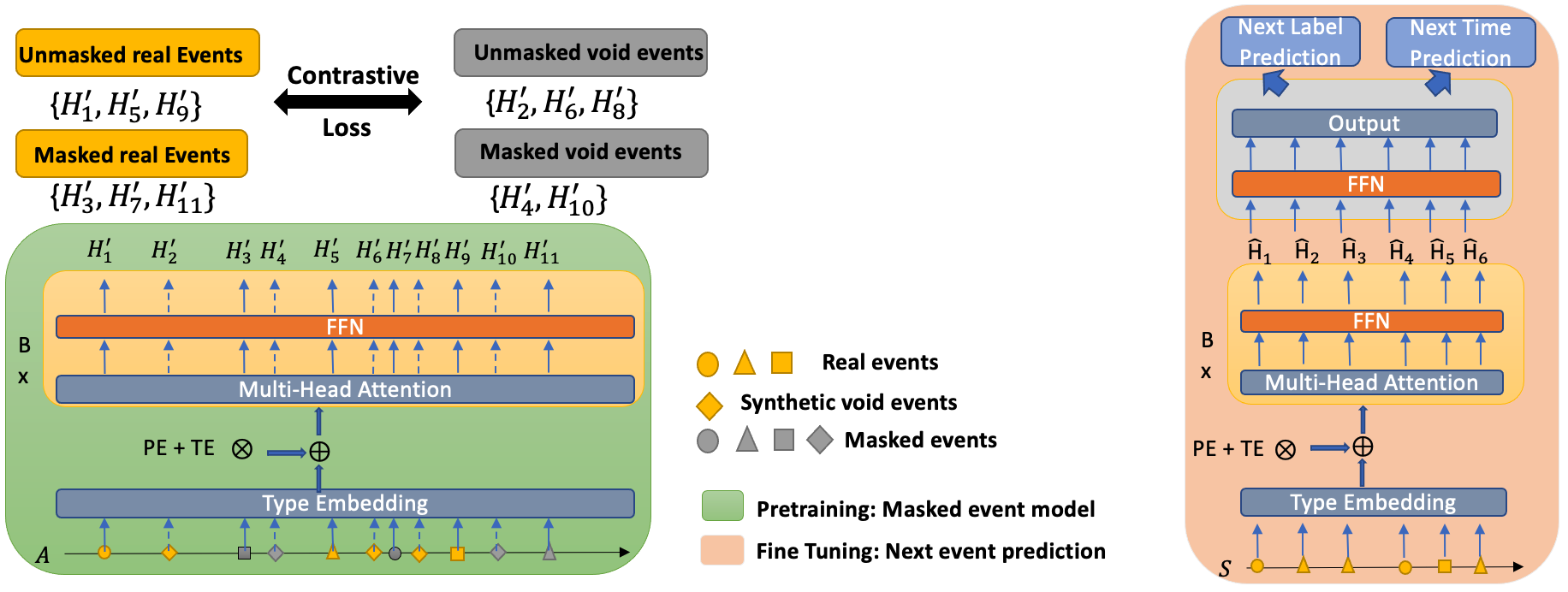}
    \caption{Pre-training and fine-tuning with Event-former. In pre-training, void events are first sampled and inserted to an event sequence $A$, randomly masked, then embedded (with combined positional and temporal encoding) and fed into a transformer network ($B$ blocks of attention). Pre-training is done by minimizing prediction error (Eq.~\ref{eqn:sample}) and contrastive loss (Eq.~\ref{eqn:contrastive}). Learned event representations for sequence $S$, $\hat{H}_i$'s are used for fine tuning with a small feed forward neural network by minimizing prediction loss  (Eq.~\ref{eqn:fine}). }% \textcolor{red}{PLEASE CHECK MY EDITS!} }
    \label{fig:mod}
\end{figure*}

We introduce a self-supervised modeling paradigm with a transformer-based architecture for multivariate event streams that we refer to as \emph{Event-former}. We note that our focus is on point process models, and to the best of our knowledge, there is no other work on self-supervised learning or transfer learning for temporal event sequences yet. 
% where self-supervision is leveraged for downstream tasks such as predicting the time as well as label of the next event, given any history. 
%Our proposed paradigm differs from current standard practices of training (neural) MTPP models, namely log-likelihood maximization via neural conditional intensity functions, which rely solely on supervised training from a single large event dataset. Instead, taking an approach that is analogous to foundation models such as BERT \citep{devlin2018bert}, we introduce novel pre-training and fine-tuning tasks for event data in continuous time.
In particular, we present a novel pretext training task specific to event data to learn a suitable representation for event streams. Such a representation can then be used by a small feed forward network for fine-tuning on a sequential next event prediction task. 
A high-level figurative scheme is shown in Figure \ref{fig:mod}; we provide details in the subsections to follow.

%Our pre-training device dubbed masked event model (MEM) differs from others in natural language process due to event happening in continuous time where non-events which we call void times or epochs in-between events are meaningful in our setting. Such void times indicate no events occur within any two consecutive positions in the sequences governed by survival dynamics. An independent Bernoulli model of masking on the event epochs only will be insufficient to capture such dynamics and consequently yield compromised performance. Nor will a geometric mask proposed for time series suffice to mitigate the performance loss. MEM is unique to event streams as the tuple $(t_i,y_i)$ are two characteristics of an event epoch and simultaneously occurring. We define MEM in the following for clarification:

%\begin{definition} 
%A masked event model is a pretraining task where some event epochs $(t_i,y_i)$'s are randomly masked for prediction for a given an event sequence ${(t_i,y_i)}_{i=1}^n$.
%\end{definition} \label{def}
%The application of void epochs (also named fake epochs) shows boosted performance in a previous study \citep{gao2020multi}. 
%Our model then leverages the learned representation of target event sequences through a pretrained model by MEM and fine tuned for next event time prediction and label prediction. 

% \subsection{Can we pretrain MTPP data as time series?} No. i think about our foundation model for marked point processes and looking at the transformer for multivariate time series (tst), i ask what is it that differentiates our approach (or what is novel) from tst, can we just treat as bivariate time series (one for time and the other for label)? If our goal is just for prediction, there is probably not that much difference; but for other tasks such as anomaly detection (we want to see if there is an actual event in the void epoch or in between 2 events) there is a difference-- at its heart, our process is continuous-time, and the void epoch is meaningful in our setting. Thus we may design a pretext task that reflects this aspect.  
Our proposed pre-training paradigm from Figure~\ref{fig:mod} (left) involves three major aspects that distinguish it from prior work: 1) injecting void events (which are formalized in the next paragraph) to improve the representation learning of event dynamics in continuous time, 2) an effective masking strategy that uses both positional and temporal encoding on the above augmented event stream with void events, and 3) forcing the attention mechanism to adhere to the temporal order of events. We explain each aspect below before prescribing the full pre-training scheme, followed by a brief explanation of the fine-tuning procedure as depicted in Figure~\ref{fig:mod} (right). 

\subsection{Void Events in Transformers}
% \paragraph{Void events in Transformers.} 
Recall that an observed event stream is of the form $S = \{(t_i,y_i)\}_{i=1}^n$ where $t_i$ and $y_i$ are the $i^{th}$ event's time stamp and label, respectively.
We consider a modified stream where we inject a predetermined number of \emph{void events} involving epochs where no event occurs; these are of the form $(t'_i, \text{null})$ where `$\text{null}$' is a new label signifying absence of an event occurrence. The modified stream is denoted $S' = \{t'_i, y'_i\}_{i=1}^{n'}$ where $S \subset S'$ and $y'_i \in \{ \mathbb{L} \cup \text{null} \}$ $\forall i$. 
The role of the void events is to provide additional information about the dynamics of the continuous-time process by explicitly indicating that no event occurs within two consecutively observed events. 

\paragraph{Void Events as Fake Epochs.} Explicitly specifying selected epochs where events do not happen has been used previously in some related work; see for instance the notion of `fake epochs' in \citep{gao2020multi}, which was originally developed for RNNs and helped boost the performance of a neural point process on a model fitting task. 
The main idea is that in point process models, the inter-event duration between two successive events is just as important as the event epochs themselves. This is seen from the integral terms in the conditional intensity based log-likelihood expression for event streams (see Eq.~\ref{eqn:ll}). Just as the internal hidden state of a recurrent neural network (such as an LSTM) only changes in discrete steps upon seeing the next token, the transformer based representations too behave in a similar manner. In reality, the conditional intensity rates can evolve continuously. Introducing the void epochs in the inter-event void space provides a convenient way to force the evolution of the transformer based internal representations inside the inter-event interval, and this in turn leads to improvement in both the pre-training and the fine-tuning steps for next event prediction related tasks.
% \textcolor{blue}{[Comment: Should we bring back some discussion about the benefits of fake epoch(s) vs. other approaches that make assumptions about the intensity decay?]}
We thereby address an inherent shortcoming of transformers for event datasets in a non-parametric manner, i.e. without the need of specific parametric or process assumptions such as in the transformer Hawkes process~\citep{zuo2020transformer}. However, to use void events in transformers requires further adaptation, particularly during the training process where masking is also used. %Next we present an effective approach for masking with void event labels.

\paragraph{Void Events as Synthetic Noises.} %A second view of void events is to treat them as synthetic noise manually injected into the sequence. Such representations should be inherently different from its real counterparts. These lead to our novel contrastive pretraining scheme for \textit{Event-former}: average representation of masked real epochs should be similar to their unmasked average real instances;  representation of masked noises should be close to their unmasked synthetic counterparts. In addition, the former representation should differ from the latter representation. Although contrastive learning and pre-training are commonly adopted in literature for time series, for example \cite{eldele2021time}; our approach employs the uniqueness of continuous time dynamics. We discuss its asscoiate loss in the following section. 
An alternative perspective on void events is to view them as artificially created noises. %that is manually inserted into the sequence. 
As a result, these synthetic events should have distinct characteristics compared to their real counterparts. This idea motivates our novel contrastive pre-training scheme for the \textit{Event-former} model. Specifically, we aim to ensure that the average representation of masked real epochs is similar to their unmasked counterparts, while the representation of masked noises is close to their unmasked synthetic counterparts. Furthermore, we expect the former representation to differ from the latter representation. \textcolor{black}{This aspect of contrastive learning in point processes largely distinguishes our approach from previous models such as Initiator \cite{guo2018initiator} and NCE-MPP \cite{mei2020noise}.} Although contrastive learning and pre-training are commonly used for time series in the literature \textcolor{black}{such as \cite{zhang2022self}}, our approach leverages the unique dynamics of continuous time. We elaborate on the associated loss function in the following section.

\subsection{Masking Strategy \& Input Encoding}

We consider a  masked event model (MEM) pre-training device specialized for point processes that operates on the modified event stream $S'$, where some events $(t'_i,y'_i)$ are randomly masked for the task of prediction given history. %Our masking approach is broadly similar to past work but specialized to our model. 
When an event epoch in the above expanded event stream $S'$ is masked, its time stamp is replaced with the value zero and its label is replaced with the value  \textit{[MASK]}. Further, for the choice of which tokens get masked, our model admits both the independent strategy used in BERT~\citep{devlin2018bert} as well as the serially correlated temporal strategy used in time series \citep{zerveas2021transformer}. An ablation study shown later in Table~\ref{tab:abl} indicates that either of these strategies works well when combined with the proposed MEM model, and leads to improvements through transfer learning in both MSE and accuracy for predicting the next event time and label respectively. We also note that the results are worse without the proposed MEM model's expansion of the event stream, i.e. without the injection of void event epochs. 
Our model differs from existing literature on masking in that both actual events and void events are admitted as candidates for masking. This combined approach leads to improved representations in experimental evaluation. The MEM model based representations are able to implicitly learn about both the event arrival rates due to masked learning with real epochs, as well as the inter-event empty spaces due to masked learning with void epochs. 
%The application of void events (also named fake epochs) shows boosted performance in a previous study \citep{gao2020multi}.

% [maybe describe the inadequate masking first, and then explain our method of addressing it?]

%due to event happening in continuous time where non-events which we call void times or epochs in-between events are meaningful in our setting. Such void times indicate no events occur within any two consecutive positions in the sequences governed by survival dynamics. 

%\begin{definition} 
%A masked event model (MEM) is a pre-training model for the prediction task where some events $(t'_i,y'_i)$ are randomly masked for prediction using history, for a given event stream $S'$.
%\end{definition} \label{def}

%[COMMENT: Edit the following and incorporate as suitable.]
%An independent Bernoulli model of masking on the event epochs only will be insufficient to capture such dynamics and consequently yield compromised performance. Nor will a geometric mask \citep{zerveas2021transformer} proposed for time series suffice to mitigate the performance loss. MEM is unique to event streams as the tuple $(t_i,y_i)$ are two characteristics of an event epoch and simultaneously occurring. 

%\paragraph{Combined position and temporal encoding.}

In addition to the choice of masking strategy in transformer models, one also needs an encoding for the position information in the input sequence so that the uniqueness of each location is retained to some extent.  Traditional positional encoding (PE) \citep{vaswani2017attention} used in transformers is not sufficient by itself for event stream data because events are associated with irregular time stamps, unlike natural language sequences. Similarly, temporal encoding (TE), such as proposed in prior work~\citep{zuo2020transformer}, also proves inadequate by itself in our setting because our masking strategy replaces the time stamps of masked events with zero. Note that this would render indistinguishable any two distinct events  (i.e. with distinct time stamps) of the same event type in the input event stream. As seen in Figure 1 in the Appendix, using TE alone leads to early plateauing of the loss function, and this is often a telltale signature of poor end-task performance. To address this issue, we propose the combined encoding strategy of using PE and TE together. We also show that the combined encoding strategy preserves the universal approximation results of standard transformers.

\begin{proposition} 
Transformers with combined PE and TE are universal approximators for any continuous sequence-to-sequence function with compact domain, i.e. they approximate any continuous functions f: $\mathbf{X}$ $\rightarrow$ $\mathbf{H}$ with $\epsilon$ error %with .respect to 
w.r.t
$p$-norm where $1 \le p < \infty$ and $\mathbf{X
}, \mathbf{H} \in \mathbb{R}^{d \times L}$.
\end{proposition} \label{thm:pete}

Please refer to the Appendix for a proof of the above result. Early work~\citep{yun2019transformers} establishes that transformers with PE are universal approximators for any continuous sequence-to-sequence function with compact support (Theorem 3 in the paper) and is applicable to language sequences. The aforementioned result however applies uniquely to event streams. More importantly, it separates two distinct event epoch encodings to (potentially) distinguish representations and establishes the predictability and learnability of a transformer model (with a certain structure) for the MEM. % Next we describe the temporal causal attention mechanism that uses only past historical information.
%\paragraph{Enforcing temporally ordered event epochs.}
\subsection{Temporal Lower Triangular Attention}
While masked language models such as BERT leverage contextual information from both prior and post tokens of interest, here we only consider prior tokens. This is because our main task of interest is event prediction given only the past, as typical in most real-world prediction problems, which 
%the temporal order of event sequences in a potential event prediction task 
prohibits us from using post token context. We apply an upper triangular mask so that a current event epoch only attends to prior events. In MEM, any representation in pre-training as shown in Figure \ref{fig:mod} (left) for a masked event (regardless of whether the event is observed or void) only attends to history in the past. With these pieces that define the MEM model, we next describe the pre-training and fine-tuning steps that respectively produce and exploit the self-supervised representations.

\subsection{Pre-training Scheme}

Pre-training using the MEM is conducted by first randomly injecting void events into event streams, masking some of the events, and then computing a self-supervised loss determined by predicting the masked events. In this fashion, the MEM is trained to not only predict the time and label of observed events, but also try to be as accurate as possible at determining when events do not happen.
In the most general setting, suppose that the sequence of time stamps for void events, denoted 
$\tau$, is randomly generated from some distribution $\mathbb{P}$. A special case of this random injection is when exactly 1 void event is uniformly generated between each pair of consecutive events in $S$ to create modified event stream $S'$.
After randomly selecting a pre-determined percentage of events to mask, the loss for the self-supervised prediction task can be computed as:
\begin{equation}
  \mathcal{L} = \mathbb{E}_{\tau \sim \mathbb{P}}  [\mathcal{L}_{event} (\hat{t}'_m, \hat{y}'_m;t'^*_m, y'^*_m )],  
  \label{eqn:every_interval}
\end{equation} 
where $m$ denote the indices of the randomly selected masked events, similar conceptually to \citep{devlin2018bert}. The hat and star notation for $t$ ($y$) refer to the model's predicted time (label) and the ground truth time (label), respectively.
Note that Eq. \ref{eqn:every_interval} will in practice be challenging to optimize, due to the stochastic objective and additional computation complexity from sampling and inserting void events between every two consecutive events in every event stream in a batch when performing stochastic gradient descent. The time complexity for such an insertion during training is $\mathcal{O}(KL)$ where $K$ is the number of event streams and $L$ is the maximum length by merging the two sorted lists. 

%[COMMENT: The following needs to be clearer.]

To reduce the computational cost and improve efficiency, we propose a practical solution by adopting a simpler but just as effective sampling strategy for void events. Specifically, we only sample void events once from the original dataset as an approximation and then merge as a pre-processing step. Thus no additional computing cost occurs during training. Let $N_M$ be the total of number of masked event epochs. We use the following to measure the prediction loss for each masked event, whether it is observed or void: 
\begin{equation}
\begin{split}
    \mathcal{L}_{pred} = \frac{1}{N_M} \sum_i^{N_M} \text{CE}(\text{softmax}(  \mathbf{H}'_m \mathbf{W}^y+\mathbf{b}^y)_{i,:\mathbf{}}, y_{m,i}^{'*}) \\
    + \gamma \text{MSE}(( \mathbf{H}'_m \mathbf{w}^t+b^t)_{i}, t_{m,i}^{'*}),
     \label{eqn:sample}
\end{split}
\end{equation}
where  $\mathbf{H}'_m$ is the masked high level representation from the transformer model of a modified stream and $\mathbf{W}^y \in \mathbb{R}^{d \times M}$, $\mathbf{b}^y \in \mathbb{R}^{M}$, $\mathbf{w}^t  \in \mathbb{R}^{d} $ and $b^t \in \mathbb{R}$ are trainable weights and biases for label prediction cross entropy (CE) loss and time prediction mean square error (MSE). In addition, index $i$ in the above equation implies a general instance of masked event epoch and $i,:$ corresponds to the $i^{th}$ row of the output matrix. We use $\gamma$ as the trade-off between the two loss terms. In addition, to improve quality of the learned representation, we enforced an additional contrastive loss in our pretraining. Let $\Tilde{\mathbf{H}}_{m,r}^{'}$, $\Tilde{\mathbf{H}}_{m,v}^{'}$ ,$\Tilde{\mathbf{H}}_{u,r}^{'}$, $\Tilde{\mathbf{H}}_{u,v}^{'}$  be the averaged masked real, averaged masked void, averaged un-masked real and averaged un-masked void representation, respectively, calculated from an modified event sequence. We propose the following contrastive loss:
\small
\begin{equation}
% \begin{split}
     \mathcal{L}_{contr} = - \text{log} \frac{\text{exp}((\text{sim}(\Tilde{\mathbf{H}}_{m,r}'  , \Tilde{\mathbf{H}}_{u,r}' )
+\text{sim}(\Tilde{\mathbf{H}}_{m,v}'  , \Tilde{\mathbf{H}}_{u,v}'))/\omega)}{\text{exp}((\text{sim}(\Tilde{\mathbf{H}}_{m,r}'  , \Tilde{\mathbf{H}}_{m,v}' ) + \text{sim}(\Tilde{\mathbf{H}}_{u,r}'  , \Tilde{\mathbf{H}}_{u,v}' ))  /\omega) },
\label{eqn:contrastive}
% \end{split}
\end{equation}
where sim(u,v) = $\frac{u ^\top v}{||u|| ||v||}$ denotes the dot product and $\omega$ is a temperature parameter, common in contrastive loss literature \cite{jaiswal2020survey}. Overall, we optimize $\mathcal{L}_{event} = \mathcal{L}_{pred} + \lambda \mathcal{L}_{contr}$ where $\lambda$ trades off the two losses. In practice, we use only one hidden layer for masked event prediction; we avoid using deep feed forward networks to force the transformer model to learn a high quality representation $\mathbf{H}'$ so that it facilitates the fine-tuning process for downstream tasks. A more detailed construction is included as Algorithm 1.

\subsection{Fine-tuning}
% [Comment: im not clear on the difference between S and each observed event. ]
After pre-training, MEM can then be applied to model any new event sequence $S$ and be further fine-tuned to obtain a better representation $\hat{\mathbf{H}}_i$. Note that during fine-tuning, we do not include void events, which simplifies the training steps and is compatible with any existing approach. Each learned representation is then fed into a small feed forward neural network for downstream tasks involving event prediction. In other words, our model fine-tunes by consuming each individual %\textcolor{blue}{[is it iid here? Xiao:  representations $\hat{H}_i$ behaves as if "iid".  ] } 
event representation $\hat{\mathbf{H}}_i$ and predicting the next label $y_{i+1}
$ as well as time $t_{i+1}$. The power of this approach is primarily through the conversion of sequential prediction into tabular regression and classification. For an event dataset with $K$ event streams, each with length $n_k$, the loss in fine tuning step $\mathcal{L}_{pred}$ is the following:
\begin{equation}
\begin{split}
    \mathcal{L}_{finetune} =  \sum_{l=1}^K \sum_{i=1}^{n_k} \text{CE}(\text{softmax}(MLP(\hat{\mathbf{H}}_i^l)),y_{i+1}^l) \\
    + \alpha \text{MSE}(MLP(\hat{\mathbf{H}}_i^l), t_{i+1}^l),
\end{split}
\label{eqn:fine}
\end{equation}
where $\alpha$ is a similar trade-off between cross entropy and mean square error. Regression and classification share the same multi-layer perceptron (MLP) for computational efficiency in our setting.

\section{Experiments}

% \subsection{Baselines}

% We use the following baselines for experiments. To focus attention on the potential benefits of self-supervision rather than the choice of neural architecture, we replace non-transformer architectures in baselines with a suitable counterpart transformer:
% \textbf{Recurrent Marked Temporal Point Process} \citep{du2016recurrent} and \textbf{Event Recurrent Point Process} \citep{xiao2017modeling}: We replace the RNNs in both originally proposed models with transformers. We note that the original implementation \footnote{https://github.com/woshiyyya/ERPP-RMTPP} only predicts the very last event $(t_n,y_n)$ given prior events; for a fair comparison, we therefore modify the code to evaluate next inter-event time $d_i$ where $d_i = t_i-t_{i-1}$ for $i \in \{2,3,...n\}$; \textbf{Lognormal Mixture} \citep{shchur2019intensity}: We replace the RNN with a transformer and take the expectation of the learned mixture model for next inter-event prediction; \textbf{Transformer Hawkes Process} \footnote{https://github.com/SimiaoZuo/Transformer-Hawkes-Process} \citep{zuo2020transformer}: This model is representative of the current state-of-the-art for event sequence modeling. It already involves a transformer architecture and therefore does not need any modification. Furthermore, it is worth noting that THP is equivalent to the model that fine-tunes all parameters on the target domain without pre-training. 

\paragraph{Baselines.} For our experiments, we establish the following baselines. To highlight the advantages of self-supervision instead of the neural architecture selection, we substitute non-transformer architectures in the baselines with a suitable transformer alternative. Specifically, we replace the RNNs in the \textbf{Recurrent Marked Temporal Point Process} \citep{du2016recurrent} and the \textbf{Event Recurrent Point Process} \citep{xiao2017modeling} models with transformers. It is important to note that the original implementation of both models only predicts the final event $(t_n, y_n)$ given previous events. To ensure a fair comparison, we modify the code to evaluate the next inter-event time $d_i$, where $d_i = t_i-t_{i-1}$ for $i \in \{2,3,...n\}$. In the \textbf{Lognormal Mixture} \citep{shchur2019intensity} model, we substitute the RNN with a transformer and calculate the expectation of the learned mixture model for the prediction of the next inter-event. The \textbf{Transformer Hawkes Process} (THP) \footnote{https://github.com/SimiaoZuo/Transformer-Hawkes-Process} \citep{zuo2020transformer}, which represents the current state-of-the-art in event sequence modeling, already employs a transformer architecture and thus requires no modification. It is also noteworthy that THP is equivalent to the model that fine-tunes all parameters on the target domain without pre-training.
%In summary we compare our approach with the above transformed based models (T-RMTPP, T-ERPP, T-LNM, and THP respectively). 
We use the following acronyms for the aforementioned models, where the prefix `T-' clarifies that some of these are transformer-based extensions: T-RMTPP, T-ERPP, T-LNM and THP. \textcolor{black}{We also include two noise contrastive models for temporal point processes in our baselines: \textbf{Initiator}~\cite{guo2018initiator} and \textbf{NCE-MPP} ~\cite{mei2020noise}, leveraging standard Pytorch implementations \footnote{https://github.com/hongyuanmei/nce-mpp}}. 
Following \citep{zuo2020transformer}, we evaluate model performance on next event time prediction with root mean square error (rmse) and on next event label prediction with accuracy(acc). Experimental code for model training and evaluations are included in the supplementary material.

\paragraph{Pretraining.} \textcolor{black}{Our pretraining model is based on the code adaptation from \cite{zuo2020transformer}. 
The entire procedure is detailed in Algorithm 1 where a combined loss $\mathcal{L}_{event} = \mathcal{L}_{predict} + \lambda \mathcal{L}_{contr}$ is minimized for each batch where $\mathcal{L}_{predict}, \mathcal{L}_{contr}$ are from equation \ref{eqn:sample} and \ref{eqn:contrastive} respectively. We first randomly sample void event instances in-between every consecutive real events and then insert such void events. Then for each augment sequence, we randomly mask 15\% of the events and split into train and dev subsets of sequences. We train our model using stochastic gradient descent and employ the Adam optimizer \cite{kingma2014adam} for optimization. The standard transformer architecture employed for pretraining is characterized by specific parameters. It comprises a multi-head self-attention module comprising 4 blocks, with a post-attention value vector dimension of 512, employing 4 attention heads. The hidden layer within the feed-forward neural network has a dimension of 1024, while both the value and key vectors are of dimension 512. Additionally, a dropout rate of 0.1 is applied. The training procedure extends over 100 epochs, utilizing a learning rate of 0.0001. The outputs of the transformer model, denoted as $\mathbf{H}_i$'s, are linearly mapped to numerical values and probabilities in the regression and classification losses presented in equation \ref{eqn:sample}. All our experiments are performed on a private server with TITAN RTX GPU.} %More details can be found via https://idea.rpi.edu/IDEA\_Cluster\_Access .

\begin{figure} [t]
    \centering
    \includegraphics[width=0.45\textwidth]{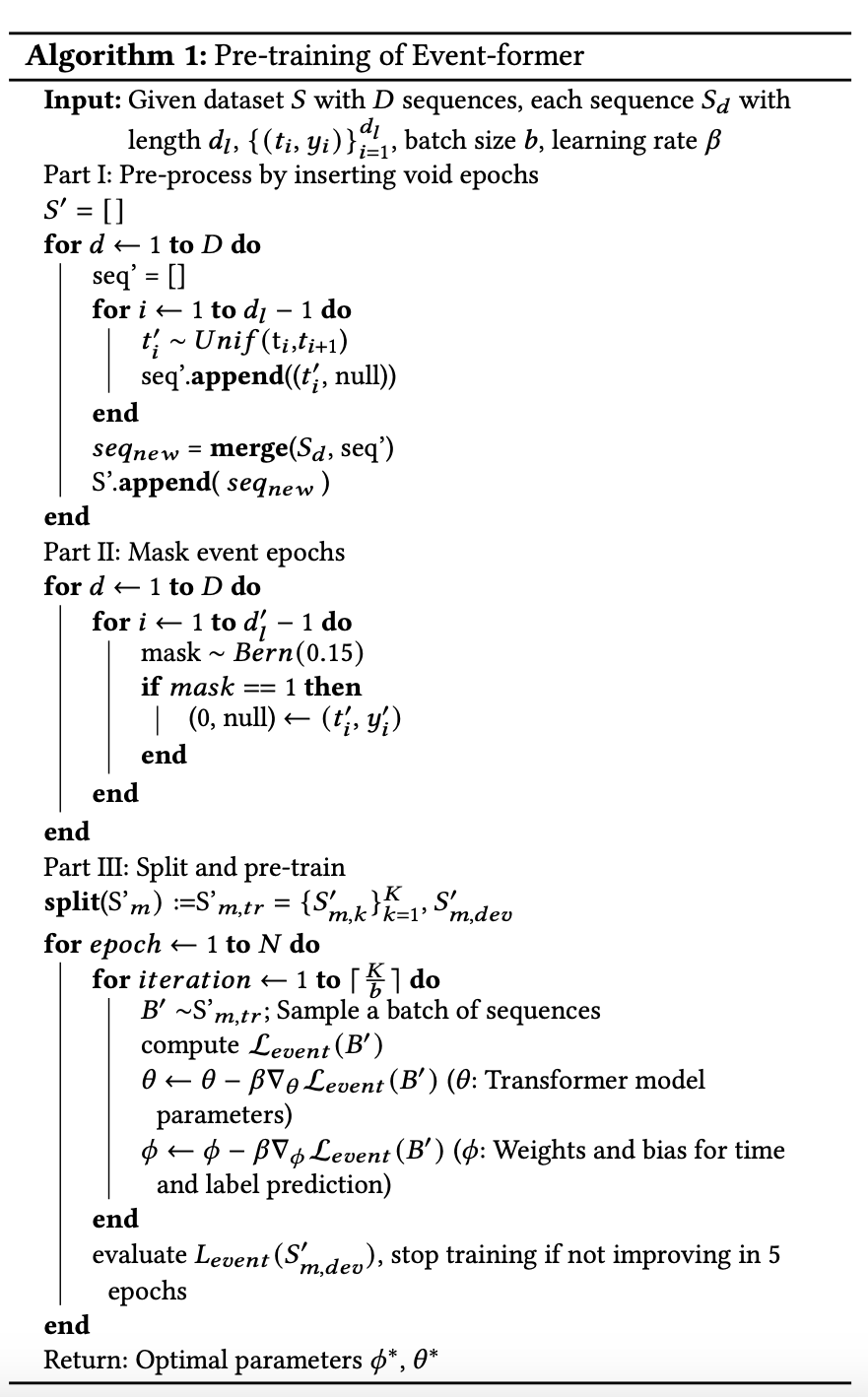}
    \label{fig:alg}
\end{figure}

% \RestyleAlgo{ruled} \label{alg:training}
% \begin{algorithm} 
% \caption{Pre-training of Event-former}
% \KwInput{Given dataset $S$ with $D$ sequences, each sequence $S_d$ with length $d_l$, $\{(t_i,y_i)\}_{i=1}^{d_l}$, batch size $b$, learning rate $\beta$ } 
% Part I: Pre-process by inserting void epochs \\
% $S' = []$ \\
% \For{$d \gets1$ \KwTo $D$ }{
%     seq' = [] \\
%     \For{$i \gets1$ \KwTo $d_l-1$ }{ 
%         $t'_i$ \sim Unif($t_i$,$t_{i+1}$) \\
%         seq'.\textbf{append}(($t'_i$, null)) 
% }
% $seq_{new}$ = \textbf{merge}($S_d$, seq') \\
% S'.\textbf{append}( $seq_{new}$ )} 
% %Part II: Mask event epochs \\
% \For{$d \gets 1$ \KwTo $D$ }{
%     \For{$i \gets1$ \KwTo $d'_l-1$ }{ 
%         mask \sim Bern(0.15) \\
%         \If{$mask ==1$} {
%         (0, null) \gets  (t'_i,y'_i) \\ }
% }
% }
% $Part III: Split and pre-train$ \\
% $\textbf{split}(S'_m)$ :=  $S'_{m,tr} = \{S'_{m,k}\}_{k=1}^{K}$, $S'_{m,dev} $ \\
% \For{$epoch\gets1$ \KwTo $N$ }{
%     \For{$iteration\gets1$ \KwTo $\lceil {\frac{K}{b}}  \rceil $  } { 
%         $B'$ \sim $S'_{m,tr}$; Sample a batch of sequences  \\
%         compute $\mathcal{L}_{event} (B')$  \\
%           \theta  \leftarrow \theta - \beta $\nabla_{\theta} \mathcal{L}_{event}(B')$  ($\theta$: Transformer model parameters) \\
%            \phi  \leftarrow \phi - \beta $\nabla_{\phi} \mathcal{L}_{event}(B')$  ($\phi$: Weights and bias for time and label prediction) \\   
%     }
%     evaluate $L_{event}(S'_{m,dev})$, stop training if not improving in 5 epochs
%     } 
% Return: Optimal parameters $\phi^{*}$, $\theta^{*}$
% \end{algorithm}

\paragraph{Fine-tuning.} \textcolor{black}{In the fine-tuning phase, a feed-forward network with three hidden layers, each having a dimension of 512, is employed. The representations extracted from the pre-trained model for both the training and development subsets are kept constant during the fine-tuning process. We opt for the Adam optimizer with a learning rate selected from the range of $\{0.001, 0.002\}$ and utilize a batch size of 32 for a total of 100 epochs. To mitigate overfitting, we incorporate early stopping, halting training if there is no improvement in the loss $\mathcal{L}_{fine-tune}$ over 10 consecutive epochs. The learning rate is adjusted based on the model's performance on the development subset of the target data.}

%For the fine-tuning process, we employ a feed-forward network consisting of three hidden layers, each with a dimension of 512. The representations of the training and development subsets are acquired from the pre-trained model and are held constant throughout the fine-tuning stage. We utilize the Adam optimizer with a learning rate chosen from the range of $\{0.001, 0.002\}$, along with a batch size of 32, for a total of 100 epochs. To prevent overfitting, we implement early stopping if there is no improvement in the loss $\mathcal{L}_{fine-tune}$ for 10 consecutive epochs. The learning rate is determined based on the model's performance on the development subset of the target data. %Furthermore, in all experiments, the value of $\alpha$ in equation 5 is fixed at 0.01.

\paragraph{Hyperparameter Selection.} \textcolor{black}{We discuss 4 important hyperparameters discussed in the model. Three in pre-training stage are the trade-off between label and time prediction loss $\gamma$ in equation \ref{eqn:sample}, the temperature parameter $\omega$ in equation \ref{eqn:contrastive}, the trade-off parameter between prediction and contrastive loss $\lambda$; the fourth in fine tuning is the trade-off parameter between time and label $\alpha$ in equation \ref{eqn:fine}. In all experiments, the hyper-parameters were selected from the best performing model on validation subset. Specifically, $\gamma$ was chosen from $\{0.01,0.1,1\}$ and empirically we found $\gamma=1$ worked well in our setting. $\omega$ could be considered as a fixed value at 1 since equation \ref{eqn:contrastive} could be simplified to a more concise form by absorbing $\omega$ into $\lambda$ in practice. $lambda$ was given the range $\{0.1,0.01,0.001,0.0001\}$ and we found $0.01$ worked best. We followed early work \cite{zuo2020transformer} and fixed $\alpha$ at 0.01 in all fine tuning experiments.}

\subsection{Synthetic Data Experiments}

We conduct experiments using synthetic data generated from two representative parametric families of multivariate temporal point processes: multivariate Hawkes processes~\citep{bacry2015hawkes} and proximal graphical event models~\citep{bhattacharjya2018proximal} (PGEM). We aim to pre-train a masked event model on a set of datasets and fine-tune the model on a different dataset for the event prediction task.  

\paragraph{Hawkes-Exp Dynamics.} We generate 400 sequences each from 10 dimensional Hawkes' process dynamics for 5 datasets (A, B, C, D, E) with different parameters and combine them to form a pre-training dataset (181175 total events) . 
We further split each into train-dev sets 75-25 and use the dev set for hyper-parameter selection. We also generate 5 folds of a dataset F with different parameters as the target; each fold contains 500 event sequences and is further split into train-dev-test 60-20-20 subsets. Final evaluation is performed on the test subsets.

\paragraph{PGEM Dynamics.} We generate 500 sequences each from the PGEM \citep{bhattacharjya2018proximal} generator with 4 datasets (A, B, C, D) of different parameters where each contains 5 event labels. We combine these to form the pre-training dataset (665883 total events). Similarly, we generate an additional 5 folds of a dataset E with different parameters as the target, each of which contains 500 event sequences. Each fold is further split into train-dev-test 60-20-20 subsets, and as before, final evaluation is performed on the test subsets. 

% a. hawkes-exp (A-E) w/ and w/o H from hawkes-exp
% b. hawkes-powerlawer (A-E) H from hawkes-powerlaw
% c. PGEM (A-E) H from PGEM
% d. learning dynamics from one generating process helps prediction for another ?

\begin{table*} [htp]
\setlength{\tabcolsep}{2.5pt}
  \caption{Next event prediction on synthetic datasets, evaluated by rmse (Time, the \textit{lower} the better) and accuracy (Type, the \textit{higher} the better). Best results  are highlighted. Standard deviations are in parentheses. }
  \centering
  \resizebox{0.80\linewidth}{!}{
  \begin{tabular}{lllllllll}
\toprule
   Dataset  & Prediction & T-RMTPP & T-ERPP  & T-LNM & THP  & \textcolor{black}{Initiator} & \textcolor{black}{NCE-MPP}  & Event-former \\
    \midrule
        Hawkes-Exp  & Time-Rmse  & 0.509(0.006) & 0.654(0.004) & 0.461(0.008) & 0.426(0.004)& 0.417(0.003)  &0.417(0.003) & \textbf{0.411(0.003)} \\
       & Type-Acc  & 0.152(0.003) &0.152(0.003) &0.172(0.004) & 0.164(0.006) & 0.088(0.001) & 0.089(0.001) & \textbf{0.179(0.003)} \\
        PGEM  &Time-Rmse  &1.068
(0.015)& 1.234(0.044) & 0.861
(0.020) & 0.777(0.010) &0.853(0.012) & 0.853(0.013)& \textbf{0.770(0.014)} \\
       &Type-Acc  &0.334 (0.005) & 0.208(0.005) & 0.336(0.013) & 0.342(0.010) &0.125(0.006) & 0.129(0.008) &\textbf{0.351(0.005)} \\
    \bottomrule
  \end{tabular}
  }
   \label{tab:syn}
\end{table*}

\paragraph{Results.} As shown in Table \ref{tab:syn}, Event-former achieves best results for predicting both the next event time and event type as compared to all baselines. For the Hawkes-Exp generated data, it boosts prediction performance by \textcolor{black}{1 $\sim$ 4\%} compared to the best baseline result. %\textcolor{red}{- For Time: the best baselines are Initiator and NCE-MPP and for type, T-LNM.};
On PGEM dataset, the increase is \textcolor{black}{1 $\sim$ 3\%} \textcolor{black}{ compared to the best baseline THP for both type and time}. The benefit of our approach along with its efficacy is its efficiency. While we only pre-train once, a typical fine-tuning procedure in this study involves a 20 fold smaller network -- %(1,056,267 trainable parameters) 
$\sim 1M$ trainable parameters
compared to say the THP model %(20,995,078 trainable parameters with its recommended setting).
with $\sim 20M$ trainable parameters at its recommended setting.
This suggests our model learns a suitable representation of the event dynamics, specially for Hawkes process.% model and its predictive performance. 

\subsection{Sequence Representations} 

\begin{figure}%
    \centering
    \subfloat[\centering Hawkes-Exp]{{\includegraphics[width=5.8cm]{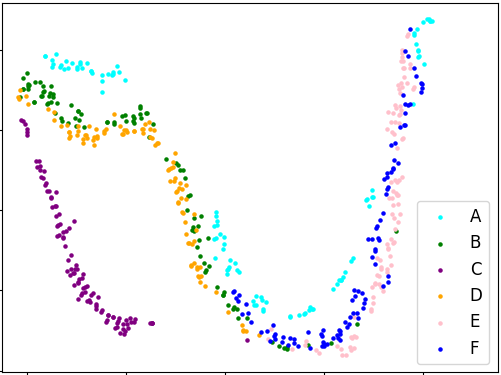} }}%
    \qquad
    \subfloat[\centering PGEM]{{\includegraphics[width=5.8cm]{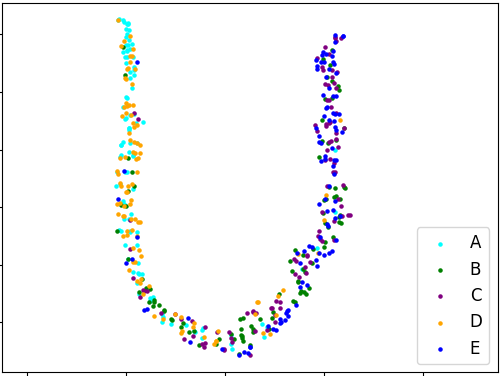} }}%
    \caption{t-SNE projection of learned representations of Hawkes-Exp and PGEM streams with pre-training on models A, B, C, D, E and F(Hawkes-Exp only) together.}%
    \label{fig:rep}%
\end{figure}

Figure \ref{fig:rep} shows the t-SNE projection of learned representations of event data generated by the above two dynamics onto the 2D plane. Each model generates a unique fragment segment that somewhat overlaps with another model. The linear and curvaceous segment pattern observed here is not uncommon for projecting time-series embedding onto a 2D plane \citep{wong2019visualizing}. \textcolor{black}{In continuous-time, event sequences share a similar learned representation.} The overlapping of the representations from pre-training data (A, B, C, D and E(for Hawkes-ExP) ) with the target data F for Hawkes and E for PGEM provides a visual explanation for how various datasets compare in terms of the learned representation. 
% For example, B and C share similar base rate and same decay rate, but different sparsity of the infectivity matrix. Details around the generating models are disclosed in Appendix. 

% \begin{wrapfigure}{R}{0.45\textwidth}
% \begin{figure} [htbp]
%     \centering
%     \includegraphics[width=0.4\textwidth]{foundation_event_models_rep_learn_hawkes.png}
%     \caption{T-SNE projection of learned representations of Hawkes-Exp streams with pre-training on models A, B, C, D and E together. }
%     \label{fig:rep}
% \end{figure}

\subsection{Real-world Applications}

\paragraph{Datasets.} %We perform transfer learning experiments on 3 pairs of real datasets from various domains, \textbf{Defi-Polygon} and \textbf{Defi-Mainnet},  \textbf{Electronics} and \textbf{Cosmetics}, and \textbf{ACLED-India} and \textbf{ACLED-Bangladesh}. 
\textcolor{black}{We conduct transfer learning experiments on three pairs of  real-world curated datasets: \textbf{Defi-Polygon} and \textbf{Defi-Mainnet} representing the realm of (decentralized) finance, \textbf{Electronics} and \textbf{Cosmetics} exemplifying the e-commerce sector, and \textbf{ACLED-India} and \textbf{ACLED-Bangladesh} depicting political conflict scenarios. A descriptive summary of the six datasets are shown in Table \ref{tab:sum}. Our approach to transfer learning and foundation models involves pretraining on a larger dataset and testing on a smaller dataset. This transfer direction is more in line with the concept of transfer learning. }

\textcolor{black}{
\textbf{Cosmetics} \footnote{https://www.kaggle.com/datasets/mkechinov/ecommerce-events-history-in-electronics-store}  and \textbf{Electronics} \footnote{https://www.kaggle.com/mkechinov/ecommerce-events-history-in-cosmetics-shop} datasets encompassed user-level online transactions in corresponding stores. These datasets were limited to December 2019 records and included four common event types: 'view,' 'cart,' 'remove-from-cart,' and 'purchase.' Transaction event sequences were filtered to be between 30 and 300 events, and timestamps were scaled to the [0,1] range. Pre-training was done on Cosmetics, followed by testing on Electronics. }

\textcolor{black}{
\textbf{Defi-Mainnet} and \textbf{Defi-Polygon} represented user-level cryptocurrency trading histories based on two different protocols \footnote{aave.com}, resulting in distinct dynamics. %Defi-Mainnet operated on the Ethereum, while Defi-Polygon used a scalable Ethereum sidechain, 
They involved six action types: 'borrow,' 'collateral,' 'deposit,' 'liquidation,' 'redeem,' and 'repay.' Timestamps were scaled into [0,1], and sequences were filtered to a length between 30 and 300 events. Pre-training was performed on Polygon, and testing on Mainnet followed. }

\textcolor{black}{
\textbf{ACLED-India}\footnote{https://www.kaggle.com/datasets/shivkumarganesh/riots-in-india-19972022-acled-dataset-50k} and \textbf{ACLED-Bangladesh}\footnote{https://www.kaggle.com/datasets/saimasharleen/acled-bangladesh} datasets recorded armed conflict events, such as riots and protests, in their respective countries. These datasets spanned different time periods (2016-2022 for India and 2010-2021 for Bangladesh) and featured various event types including 'Battles', 'Explosions/Remote violence',  'Riots',
        'Violence against civilians','Protests',and 'Strategic developments'.
% . For India, there were six event types, while Bangladesh shared four event types with India. 
Sequences shorter than 2 events and longer than 300 were filtered out, and timestamps were scaled to [0,1]. Pre-training was done on India, and testing on Bangladesh was carried out. 
}

\begin{table*} [h]
\setlength{\tabcolsep}{4pt}
  \caption{Properties of 6 real datasets.}
  \label{lab:dataset}
  \centering
  \resizebox{0.58\linewidth}{!}{
  \begin{tabular}{llllll}
    \toprule
    % \multicolumn{2}{c}{Part}                   \\
    %\cmidrule(r){1-2}
   Dataset  & \# classes  & \# seqs. & Avg. length & \# events & Data Type \\
    \midrule
        \textbf{Defi-Mainnet}   & 6  & 20539 & 32 & 654844 & Financial \\
        \textbf{Defi-Polygon}  & 6  & 33597 & 85 & 2856453 & Financial \\
        \textbf{Electronics}  & 4 & 9993 & 20 & 195726 &E-Commerce \\
       \textbf{Cosmetics} & 4  & 19301 & 39 & 752109 & E-Commerce \\
      \textbf{ACLED-India} & 6  & 111 & 17 & 1934& Political \\
       \textbf{ACLED-Bangladesh} & 4 & 97  & 17 & 1697 & Political \\
        % Cosmetics-trim & 4 & multiclass &  19669 & 25 & 1 & Real e-commerce \\
    \bottomrule
  \end{tabular}
  }
 \label{tab:sum}
\end{table*}

% The six datasets contain various number of sequences: 33,597; 20,539; 19,301; 9,993; 111; and 97, respectively. More details are shown in Table 1 in the Appendix. 

\begin{table*} [htp]
\setlength{\tabcolsep}{2.5pt}
  \caption{Next event prediction for 3 real applications. “-” means the learning method fails to converge. Relative improvement of Event-former over best baselines for each evaluation is shown.}
  \centering
   \resizebox{0.75\linewidth}{!}{
  \begin{tabular}{lccccccccl}
\toprule
   Dataset  & Prediction & T-RMTPP & T-ERPP  & T-LNM & THP  & \textcolor{black}{Initiator} & \textcolor{black}{NCE-MPP} & Event-former & Improvement \\
    \midrule
        Defi-Mainnet   
       & Time-Rmse  & 1.711 &0.989 &0.056  & 0.055 & \textbf{0.046} & \textbf{0.046} & 0.048 & 0\%  \\
       & Type-Acc  & 0.507 & 0.480 & 0.486 & 0.494& 0.230& 0.252 & \textbf{0.578} & 14\% $\uparrow$  \\
        Electronics  &Time-Rmse  &0.055& 0.984 & 0.011& 0.012  & - & - & \textbf{0.010} & 9\% $\downarrow$ \\
       &Type-Acc  &0.821 & 0.823 & 0.820& 0.809 & - & - & \textbf{0.987} & 20\% $\uparrow$ \\
        Bangladesh  &Time-Rmse  &2.200& 0.966 &0.092
& 0.101  & - & \textbf{0.060} & 0.076 & 0\% \\
       &Type-Acc  &0.673& 0.676 & 0.647 & 0.630 & - & 0.599 & \textbf{0.700} & 4\% $\uparrow$ \\
    \bottomrule
  \end{tabular}
   }
   \label{tab:real}
\end{table*}  

\begin{table*} [h!]
  \caption{ Ablation study on the effect of void events, masking type and percentage, evaluated by rmse (Time) and accuracy (Type). Standard deviations are in parentheses. Upward and downward arrows $\uparrow$ , $\downarrow$  indicate increase and decrease compared to Event-former results in Table \ref{tab:syn}; and "$-$" indicates remaining the same. }
  \centering
 \resizebox{0.76\linewidth}{!}{
  \begin{tabular}{lllllll}
\toprule
 &\multicolumn{2}{c}{No-void Injection}
 &\multicolumn{2}{c}{Geometric Mask }
 &\multicolumn{2}{c}{Mask Fraction 30\%}
\\\cmidrule(r){2-3}\cmidrule(l){4-5} \cmidrule(l){6-7} 
   Dataset & Time & Type & Time & Type  & Time & Type  \\
    \midrule
        Hawkes-Exp   & 0.411(0.003) $-$ & 0.175(0.003) $\downarrow$ & 0.411(0.003) $-$ & 0.176(0.003) $\downarrow$ & 0.411(0.003) $-$ & 0.175(0.003) $\downarrow$ \\
        PGEM    & 0.772(0.013) $\uparrow$ & 0.348(0.003) $\downarrow$ & 0.772(0.014) $\uparrow$ & 0.350(0.004) $\downarrow$ & 0.770(0.014) $-$ & 0.349(0.004) $\downarrow$ \\
    \bottomrule
  \end{tabular}
   }
      \label{tab:abl}
\end{table*}

\begin{table*} [h!]
\setlength{\tabcolsep}{2.5pt}
  \caption{Results of baselines trained with both source and target for all real datasets. }
  \centering
  \resizebox{0.78\linewidth}{!}{
  \begin{tabular}{llllllllll}
\toprule
   Dataset  & Prediction & T-RMTPP+ & T-ERPP+  & T-LNM+ & THP+  & \textcolor{black}{Initiator+} & \textcolor{black}{NCE-MPP+} & Event-former&Improvement \\
    \midrule
%         Hawkes-Exp  & Time  & 0.508
% (0.006) & 1.687(0.189) & 0.452(0.012) & 0.417(0.005) \\
%        & Type  & 0.144
% (0.004) & 0.134(0.003) & 0.156(0.005) & 0.163(0.006) \\
%         PGEM  &Time  &1.066(0.015)& 4.338(0.674) & 0.834(0.014) & 0.772(0.012)  \\
%        &Type  &0.343(0.006) & 0.347(0.006) & 0.340(0.011)& 0.341(0.008) \\
    % \bottomrule
        Defi-Mainnet   
       & Time-Rmse  & 2.276 &0.978&0.056 & 0.053 & \textbf{0.046}& \textbf{0.046} &0.048 & 0\%\\
       & Type-Acc  & 0.485 &0.491 & 0.518 & 0.427 & 0.140& 0.174 & \textbf{0.578} & 12\% $\uparrow$\\
        Electronics  &Time-Rmse  &0.014& 0.971& 0.011& 0.017 & - & - & \textbf{0.010}& 9\% $\downarrow$ \\  
       &Type-Acc  &0.826 & 0.824 & 0.811& 0.809 & - & - & \textbf{0.987}& 19\% $\uparrow$ \\
        Bangladesh  &Time-Rmse  &1.973& 0.995&0.091
& 0.133  &0.072 & \textbf{0.067} & 0.076 & 0\%  \\
       &Type-Acc  &0.688& 0.676 &0.650& 0.644 & 0.669 &0.635 &\textbf{0.700} & 4\% $\uparrow$ \\ 
    \bottomrule    
  \end{tabular}
  }
   \label{tab:synplus}
\end{table*}  
%The 3 target datasets where fine-tuning takes place are 1). \textbf{Defi-Mainnet}  2). \textbf{Electronics} and 3). \textbf{ACLED-Bangladesh}. We use additional 3 different datasets for pretraining Event-former: 1). \textbf{Defi-Polygon} 2). \textbf{Cosmetics} and 3).\textbf{ACLED-India} correspondingly. 

%\textbf{Defi-Mainnet} and \textbf{Defi-Polygon} are collected and privately curated datasets involving user-level crypto transactions from Aave website\footnote{aave.com}. Mainnet and Polygon represent two different protocols/deployments on the platform. \textbf{Electronics} and \textbf{Cosmetics} contain user-level online transactions for electronic \footnote{https://www.kaggle.com/datasets/mkechinov/ecommerce-events-history-in-electronics-store} and cosmetic products \footnote{https://www.kaggle.com/mkechinov/ecommerce-events-history-in-cosmetics-shop}. \textbf{ACLED-Bangladesh} \footnote{https://www.kaggle.com/datasets/saimasharleen/acled-bangladesh} and \textbf{ACLED-India} \footnote{https://www.kaggle.com/datasets/shivkumarganesh/riots-in-india-19972022-acled-dataset-50k} are political conflicts datasets; and each involves an actor in a sequence of conflicting actions (i.e. riots and protests) in the respective country. A descriptive summary of the 6 datasets are shown in Table \ref{tab:sum} in Appendix. 

\paragraph{Results.} 
As demonstrated by its highest accuracy and lowest rmse in Table \ref{tab:real}, 
  transfer learning with Event-former in these datasets consistently improves upon %all 
  \textcolor{black}{most} baselines, and the improvement ranges from $4\%$ to $20\%$. The most impressive improvement of 20\% is on time prediction in Electronics where using a different product appears to be sufficient to help with learning the dynamics. This likely suggests that the dynamics of real applications have noticeable shared similarities that the state-of-the-art transformer model approaches are unable to exploit. % For instance, a user may perform lots of (repeated) actions such as view an item many times, and placing it in the shopping cart, and then remove again in a few seconds. If it were some expensive electronic product, the decision of purchasing may come after months of consideration. A generative point process model may find the event of purchase after months' inactivity highly unlikely because of the governing survival dynamics for next event in the generative processes.
  In addition, it is worth noting that our transfer model works well in any datasets regarding sample sizes and number of events (see Table 1 in the Appendix for more details). 
  %increases with increasing size of  pre-training data. %As shown from Table 1 in the Appendix,  the 3 pairs of datasets  have different amounts of data; the Defi datasets and E-commerce datasets have mores samples than the ACLED.
  This is an indication that Event-former may be able to generalize well with appropriate pre-training data. \textcolor{black}{We note that although the two non-self-supervised baselines (Initiator and NCE-MPP) predict time well, they fail on type prediction.%, compared with Event-former.  %The success of self-supervised contrastive learning with time series data \cite{zhang2022self} further supports the idea that the evolution of contrastive learning paradigms will predominantly center on self-supervised pretraining.}
  The superior performance of Event-former and its time series counterpart \cite{zhang2022self} suggests that the future direction of contrastive learning paradigms may lie in the space of self-supervised pre-training.}
  %The improvement of pre-training is consistent with the amount of data available, which is another positive indication that Event-former can utilize pre-training data for better generalization.

\subsection{Ablation Experiments}

\paragraph{Ablational Masking Experiments.}
A typical deployment of MEM involves 3 components: inserted random void epochs, random selection of masks and masking fraction. Our default setting is through the use of void events and uniformly randomly selecting 15\% for masking during pre-training. We perform 3 ablation studies on the synthetic datasets: 
1) \textbf{void vs. no void events} \footnote{eq.4 reduces to negative similarity between the real masked vs. unmasked events}, 
2) \textbf{geometric vs. random mask} and 3) \textbf{mask fraction}. Ablation 1 evaluates the effect of injected random void epochs in MEM on prediction. Clearly from Table \ref{tab:abl}, we notice either a drop of type accuracy or increase of rmse (or both) for both datasets. The injection of void epochs is justified for producing competitive results in random masking. Ablation 2 compares the impact of two masking strategies: geometric and random. We employ the former from \citep{zerveas2021transformer}, along with inserted void epochs. Geometric masks produce slightly deteriorated results particularly on both simulations, suggesting masking consecutive segments may not aid in learning dynamics in the continuous-time setting. Ablation 3 compares the choice of fraction of randomly masked epochs. In general, we find a slight better performance for using 15\% masking for event prediction.

% \paragraph{Discussion on Pre-training Strategies}
% Event-former is trained to minimize the loss associated with Eq. \ref{eqn:every_interval} \& \ref{eqn:sample}. This approach can be considered as discriminative, better suitable for any downstream prediction tasks. It can however be converted to some generative point process~\citep{li2021mitigating}. Alternatively, one can choose Eq. \ref{eqn:ll} to optimize log-likelihood. We find empirically that our model benefits from choosing the former. Minimizing the latter loss in pre-training results in much inferior results as shown in Table \ref{tab:1vs23}. Since we choose to optimize with the former, we are better suited for injecting void epochs. Note that Eq. \ref{eqn:ll} naturally embeds the notion of continuous time in the conditional intensity function and the integral term of non events as well. However, pre-training with  Eq. \ref{eqn:every_interval} \& \ref{eqn:sample} effectively loses the notion of continuous time; we therefore bring it back through the injection of void epochs. 

\paragraph{Baselines Trained with Source and Target.}
To further demonstrate the power of our self-supervised paradigm, we train baseline models with \textit{both source and target datasets} and evaluate on the target test subset on real benchmarks in Table \ref{tab:synplus}.These baselines are denoted with post-fix "+".
It is worth pointing out that Event-former still outperforms these strong baselines in most datasets. Specifically, \textcolor{black}{adding the source datasets in the training of these baselines does not significantly improve their generalization in the test datasets, which further validates that the source and target datasets are of varying dynamics and thus suggests the value of homogeneous transfer in continuous time event streams. }%compared to these baselines, Event-former consistently improves by 9\%, 9\%, 16\% reduction of time prediction errors, and 12\%, 19\%, 2\% increase of type prediction accuracy on Defi-Mainnet, Electronics and Bangladesh respectively.  

\section{Conclusion}
In this work, we proposed a novel self-supervised paradigm for transfer learning in multivariate temporal point processes.  We introduced the usage of void events for transformer architectures, which was unique in continuous-time event models, and designed a corresponding masking strategy for predicting masked event epochs and void spaces in-between. We empirically demonstrated the potential of our approach using synthetic as well as various real-world datasets. In particular, improvement of prediction performance was noticeably significant on transferring tasks over many existing competitive transformer-based approaches. While this study focused on the homogeneous transfer setting, our approach could potentially be extended to other more complex transfer settings, such as out-of-domain heterogeneous transfer~\citep{9134370} with datasets that contain non-overlapping event labels. \textcolor{
black}{A plausible approach for heterogeneous transfer can be achieved via reprogramming \cite{yang2021voice2series}, which is a potential direction for future work.} 

% \bibliography{aaai24}
%%
%% The acknowledgments section is defined using the "acks" environment
%% (and NOT an unnumbered section). This ensures the proper
%% identification of the section in the article metadata, and the
%% consistent spelling of the heading.
% \begin{acks}
% To Robert, for the bagels and explaining CMYK and color spaces.
% \end{acks}

%%
%% The next two lines define the bibliography style to be used, and
%% the bibliography file.
\bibliographystyle{ACM-Reference-Format}
\bibliography{main}

%%
%% If your work has an appendix, this is the place to put it.
\appendix

\section{Synthetic Generators}
We generated datasets from Hawkes process and Proximal Graphical Event Model (PGEM). We describe the parameters used in our experiments to generate event datasets.

\paragraph{Hawkes-Exp.} We use a standard library to simulate 10-dimensional event sequences from the Hawkes dynamics with exponential decay\footnote{https://x-datainitiative.github.io/tick/}.The parameters which control the generation are baseline rate, decay coefficent, adjacency (infectivity matrix) and end time.
The four parameters for models A, B, C, D, E and F in our study are different from each other, and thus represent a variety of generating mechanisms which are implied by differing baseline rates, decaying patterns, triggering behaviors and sequence time horizons. The actual values for the 6 models used in our experiments are listed in the following for the sake of reproducibility. \\
\\
\textbf{A}. baseline = [0.1097627 , 0.14303787, 0.12055268, 0.10897664, 0.08473096,
       0.12917882, 0.08751744, 0.1783546 , 0.19273255, 0.0766883], decay = 2.5, infectivity = [[0.15037453, 0.10045448, 0.10789028, 0.17580114, 0.01349208,
        0.01654871, 0.00384014, 0.1581418 , 0.14779747, 0.16524382],
       [0.1858717 , 0.1517864 , 0.08765006, 0.14824807, 0.02246419,
        0.12154197, 0.02722749, 0.17942359, 0.0991161 , 0.07875789],
       [0.05024778, 0.14705235, 0.0866379 , 0.10796424, 0.0035688 ,
        0.11730922, 0.11625704, 0.11717598, 0.17924869, 0.12950002],
       [0.06828233, 0.08300669, 0.13250303, 0.01143879, 0.12664085,
        0.12737611, 0.03995854, 0.02448733, 0.05991018, 0.0690806 ],
       [0.10829905, 0.0833048 , 0.18772458, 0.01938165, 0.03967254,
        0.03063796, 0.12404668, 0.04810838, 0.0885677 , 0.04642443],
       [0.03019353, 0.02096386, 0.1246585 , 0.02624547, 0.03733743,
        0.07003299, 0.15593352, 0.01844271, 0.1591532 , 0.01825224],
       [0.18546165, 0.08901222, 0.18551894, 0.11487999, 0.14041038,
        0.00744305, 0.05371431, 0.02282927, 0.05624673, 0.02255028],
       [0.06039543, 0.07868212, 0.01218371, 0.13152315, 0.10761619,
        0.05040616, 0.09938195, 0.01784238, 0.10939112, 0.1765038 ],
       [0.06050668, 0.1267631 , 0.02503273, 0.13605401, 0.0549677 ,
        0.03479404, 0.11139803, 0.00381908, 0.15744288, 0.00089182],
       [0.12873957, 0.05128336, 0.13963744, 0.18275114, 0.04724637,
        0.10943116, 0.11244817, 0.10868939, 0.04237051, 0.18095826]] and end-time = 10. 
        
\textbf{B}. Baseline = [8.34044009e-02, 1.44064899e-01, 2.28749635e-05, 6.04665145e-02,
       2.93511782e-02, 1.84677190e-02, 3.72520423e-02, 6.91121454e-02,
       7.93534948e-02, 1.07763347e-01], decay = 2.5, 
      adjacency = [[0.        , 0.        , 0.03514877, 0.15096311, 0.        ,
        0.11526461, 0.07174169, 0.09604815, 0.02413487, 0.        ],
       [0.        , 0.        , 0.        , 0.        , 0.15066599,
        0.15379789, 0.01462053, 0.00671417, 0.        , 0.        ],
       [0.01690747, 0.07239546, 0.16467728, 0.        , 0.11894528,
        0.        , 0.11802102, 0.14348615, 0.00314406, 0.12896239],
       [0.17000181, 0.        , 0.        , 0.        , 0.        ,
        0.        , 0.        , 0.0504772 , 0.04947341, 0.        ],
       [0.        , 0.11670321, 0.03638242, 0.        , 0.        ,
        0.00917392, 0.        , 0.0252251 , 0.10131151, 0.1203002 ],
       [0.        , 0.07118317, 0.11937903, 0.07120436, 0.        ,
        0.        , 0.        , 0.08851807, 0.16239168, 0.10083865],
       [0.        , 0.02363421, 0.02394394, 0.1388041 , 0.06836732,
        0.02842716, 0.15945428, 0.        , 0.        , 0.12481123],
       [0.15185513, 0.        , 0.1290996 , 0.        , 0.        ,
        0.15401787, 0.        , 0.16587219, 0.11405672, 0.10687992],
       [0.        , 0.        , 0.07734744, 0.09943488, 0.07016556,
        0.04074891, 0.        , 0.        , 0.00049346, 0.10609756],
       [0.05615574, 0.09061013, 0.15230831, 0.06142066, 0.15619243,
        0.10716606, 0.00271994, 0.15978585, 0.11877677, 0.17145652]] and end-time = 40.

\textbf{C}. baseline = [0.08719898, 0.00518525, 0.1099325 , 0.08706448, 0.08407356,
       0.06606696, 0.04092973, 0.12385419, 0.05993093, 0.05336546], decay = 2.5, adjacency = [[0.        , 0.09623737, 0.        , 0.        , 0.        ,
        0.14283231, 0.        , 0.        , 0.        , 0.        ],
       [0.        , 0.        , 0.        , 0.        , 0.        ,
        0.        , 0.        , 0.        , 0.        , 0.        ],
       [0.        , 0.        , 0.        , 0.        , 0.        ,
        0.        , 0.14434229, 0.10548788, 0.        , 0.        ],
       [0.        , 0.        , 0.16178088, 0.        , 0.        ,
        0.        , 0.        , 0.        , 0.        , 0.        ],
       [0.        , 0.        , 0.        , 0.        , 0.        ,
        0.        , 0.        , 0.        , 0.        , 0.        ],
       [0.14554646, 0.        , 0.        , 0.        , 0.        ,
        0.09531432, 0.        , 0.        , 0.        , 0.        ],
       [0.        , 0.        , 0.        , 0.        , 0.        ,
        0.        , 0.        , 0.        , 0.        , 0.04899491],
       [0.        , 0.        , 0.        , 0.07048057, 0.        ,
        0.07546073, 0.        , 0.        , 0.        , 0.        ],
       [0.05697369, 0.        , 0.        , 0.        , 0.        ,
        0.        , 0.11709833, 0.        , 0.03100542, 0.        ],
       [0.        , 0.        , 0.        , 0.        , 0.        ,
        0.        , 0.04016475, 0.        , 0.10768499, 0.06297179]] and end-time = 80.
        
\textbf{D}. Baseline = [0.11015958, 0.14162956, 0.05818095, 0.10216552, 0.17858939,
       0.17925862, 0.02511706, 0.04144858, 0.01029344, 0.08816197], decay = [[8., 9., 2., 7., 3., 3., 2., 4., 6., 9.],
       [2., 9., 8., 9., 2., 1., 6., 5., 2., 6.],
       [5., 8., 7., 1., 1., 3., 5., 6., 9., 9.],
       [8., 6., 2., 2., 2., 6., 6., 8., 5., 4.],
       [1., 1., 1., 1., 3., 3., 8., 1., 6., 1.],
       [2., 5., 2., 3., 3., 5., 9., 1., 7., 1.],
       [5., 2., 6., 2., 9., 9., 8., 1., 1., 2.],
       [8., 9., 8., 5., 1., 1., 5., 4., 1., 9.],
       [3., 8., 3., 2., 4., 3., 5., 2., 3., 3.],
       [8., 4., 5., 2., 7., 8., 2., 1., 1., 6.]], adjacency = [[0.        , 0.14343731, 0.        , 0.11247478, 0.        ,
        0.        , 0.        , 0.09416725, 0.        , 0.        ],
       [0.        , 0.        , 0.        , 0.        , 0.        ,
        0.        , 0.        , 0.15805746, 0.        , 0.08262718],
       [0.        , 0.        , 0.03018515, 0.        , 0.        ,
        0.        , 0.        , 0.        , 0.        , 0.        ],
       [0.        , 0.11090719, 0.08158544, 0.        , 0.11109462,
        0.        , 0.        , 0.        , 0.        , 0.        ],
       [0.        , 0.        , 0.14264684, 0.        , 0.        ,
        0.11786607, 0.        , 0.        , 0.01101593, 0.        ],
       [0.04954495, 0.        , 0.        , 0.        , 0.        ,
        0.12385743, 0.        , 0.        , 0.02375575, 0.05345351],
       [0.        , 0.14941748, 0.02618691, 0.        , 0.13608937,
        0.        , 0.06263167, 0.        , 0.04097688, 0.14101171],
       [0.        , 0.11902986, 0.        , 0.04889382, 0.        ,
        0.        , 0.01569298, 0.03678315, 0.        , 0.        ],
       [0.05359555, 0.        , 0.        , 0.09188512, 0.        ,
        0.14255311, 0.        , 0.        , 0.        , 0.        ],
       [0.12792415, 0.05843994, 0.16156482, 0.11931973, 0.        ,
        0.00774966, 0.00947755, 0.        , 0.        , 0.        ]], and end-time = 50.

\textbf{E}. Baseline = [0.2, 0.2, 0.2, 0.2, 0.2, 0.2, 0.2, 0.2, 0.2, 0.2], decay = [[9., 4., 9., 9., 1., 6., 4., 6., 8., 7.],
       [1., 5., 8., 9., 2., 7., 3., 3., 2., 4.],
       [6., 9., 2., 9., 8., 9., 2., 1., 6., 5.],
       [2., 6., 5., 8., 7., 1., 1., 3., 5., 6.],
       [9., 9., 8., 6., 2., 2., 2., 6., 6., 8.],
       [5., 4., 1., 1., 1., 1., 3., 3., 8., 1.],
       [6., 1., 2., 5., 2., 3., 3., 5., 9., 1.],
       [7., 1., 5., 2., 6., 2., 9., 9., 8., 1.],
       [1., 2., 8., 9., 8., 5., 1., 1., 5., 4.],
       [1., 9., 3., 8., 3., 2., 4., 3., 5., 2.]],
      adjacency= [[0.02186539, 0.09356695, 0.        , 0.16101355, 0.11527002,
        0.09149395, 0.        , 0.        , 0.15672219, 0.        ],
       [0.        , 0.        , 0.14241135, 0.        , 0.11167029,
        0.        , 0.        , 0.        , 0.0934937 , 0.        ],
       [0.        , 0.        , 0.        , 0.        , 0.        ,
        0.        , 0.        , 0.        , 0.15692693, 0.        ],
       [0.08203618, 0.        , 0.        , 0.02996925, 0.        ,
        0.        , 0.        , 0.        , 0.        , 0.        ],
       [0.        , 0.        , 0.11011391, 0.08100189, 0.        ,
        0.11029999, 0.        , 0.        , 0.        , 0.        ],
       [0.        , 0.        , 0.        , 0.14162654, 0.        ,
        0.        , 0.11702301, 0.        , 0.        , 0.01093714],
       [0.        , 0.04919057, 0.        , 0.        , 0.        ,
        0.        , 0.12297152, 0.        , 0.        , 0.02358583],
       [0.05307117, 0.        , 0.14834875, 0.02599961, 0.        ,
        0.13511597, 0.        , 0.06218368, 0.        , 0.04068378],
       [0.1400031 , 0.        , 0.11817848, 0.        , 0.0485441 ,
        0.        , 0.        , 0.01558073, 0.03652006, 0.        ],
       [0.        , 0.05321219, 0.        , 0.        , 0.0912279 ,
        0.        , 0.14153347, 0.        , 0.        , 0.        ]] and end-time = 50. 
        
\textbf{F}. Baseline = [0.21736198, 0.11134775, 0.16980704, 0.33791045, 0.00188754,
       0.04862765, 0.26829963, 0.3303411 , 0.05468264, 0.23003733], decay = [[5., 1., 7., 3., 5., 2., 6., 4., 5., 5.],
       [4., 8., 2., 2., 8., 8., 1., 3., 4., 3.],
       [6., 9., 2., 1., 8., 7., 3., 1., 9., 3.],
       [6., 2., 9., 2., 6., 5., 3., 9., 4., 6.],
       [1., 4., 7., 4., 5., 8., 7., 4., 1., 5.],
       [5., 6., 8., 7., 7., 3., 5., 3., 8., 2.],
       [7., 7., 1., 8., 3., 4., 6., 5., 3., 5.],
       [4., 8., 1., 1., 6., 7., 7., 6., 7., 5.],
       [8., 4., 3., 4., 9., 8., 2., 6., 4., 1.],
       [7., 3., 4., 5., 9., 9., 6., 3., 8., 6.]], adjacency = [[0.15693854, 0.04896059, 0.0400508 , 0.        , 0.        ,
        0.13021228, 0.        , 0.10699903, 0.        , 0.15329807],
       [0.0784283 , 0.        , 0.        , 0.        , 0.        ,
        0.        , 0.00310706, 0.0090892 , 0.07758874, 0.        ],
       [0.01672489, 0.        , 0.        , 0.07851303, 0.        ,
        0.        , 0.12848331, 0.08859293, 0.        , 0.        ],
       [0.09984995, 0.        , 0.        , 0.10541925, 0.        ,
        0.08032527, 0.        , 0.        , 0.        , 0.        ],
       [0.14642469, 0.06629365, 0.        , 0.        , 0.        ,
        0.        , 0.11891738, 0.04166225, 0.09808829, 0.17638655],
       [0.00976324, 0.1100343 , 0.02003261, 0.        , 0.        ,
        0.05993539, 0.09739541, 0.        , 0.        , 0.        ],
       [0.        , 0.        , 0.        , 0.04672133, 0.16916   ,
        0.        , 0.17341419, 0.12078975, 0.14441602, 0.        ],
       [0.        , 0.17305542, 0.06927975, 0.        , 0.03408974,
        0.        , 0.08457162, 0.03787486, 0.10863292, 0.        ],
       [0.07186225, 0.05760593, 0.        , 0.        , 0.08042031,
        0.04403479, 0.1033595 , 0.17046747, 0.        , 0.05083523],
       [0.        , 0.10029222, 0.        , 0.1022067 , 0.        ,
        0.        , 0.0588527 , 0.        , 0.03530513, 0.        ]], and end-time = 40. 

\paragraph{PGEM.} We implement the PGEM generator by previous work \cite{bhattacharjya2018proximal} and generate 5-dimensional event datasets governed by the PGEM dynamics, which is signified by its piecewise conditional intensity function for each event type at a given time instance of a parental configuration observed at a particular window. The parameters are the conditional intensity (lambdas) for each event type given parental states, parental configuration (parents), windows for each parental state (windows) and end time which is 100 across all models. The parameters for models A, B, C, D and E  are listed in the following.  We point out to the readers that the 5 models share different graphs, which is implied by the parents for each event type (node).

\textbf{A} has the following parameters: {'parents': {'A': [],
    'B': [],
    'C': ['B'],
    'D': ['A', 'B'],
    'E': ['C']},
   'windows': {'A': [], 'B': [], 'C': [15], 'D': [15, 30], 'E': [15]},
   'lambdas': {'A': {(): 0.2},
    'B': {(): 0.05},
    'C': {(0,): 0.2, (1,): 0.3},
    'D': {(0, 0): 0.1, (0, 1): 0.05, (1, 0): 0.3, (1, 1): 0.2},
    'E': {(0,): 0.1, (1,): 0.3}}}

\textbf{B} has the follow parameters: {'parents': {'A': ['B'],
    'B': ['B'],
    'C': ['B'],
    'D': ['A'],
    'E': ['C']},
   'windows': {'A': [15], 'B': [30], 'C': [15], 'D': [30], 'E': [30]},
   'lambdas': {'A': {(0,): 0.3, (1,): 0.2},
    'B': {(0,): 0.2, (1,): 0.4},
    'C': {(0,): 0.4, (1,): 0.1},
    'D': {(0,): 0.05, (1,): 0.2},
    'E': {(0,): 0.1, (1,): 0.3}}}.

\textbf{C} has the follow parameters:
{'parents': {'A': ['B', 'D'],
    'B': [],
    'C': ['B', 'E'],
    'D': ['B'],
    'E': ['B']},
   'windows': {'A': [15, 30], 'B': [], 'C': [15, 30], 'D': [30], 'E': [30]},
   'lambdas': {'A': {(0, 0): 0.1, (0, 1): 0.05, (1, 0): 0.3, (1, 1): 0.2},
    'B': {(): 0.2},
    'C': {(0, 0): 0.2, (0, 1): 0.05, (1, 0): 0.4, (1, 1): 0.3},
    'D': {(0,): 0.1, (1,): 0.2},
    'E': {(0,): 0.1, (1,): 0.4}}}.
    
\textbf{D} has the follow parameters: 
{'parents': {'A': ['B'],
    'B': ['C'],
    'C': ['A'],
    'D': ['A', 'B'],
    'E': ['B', 'C']},
   'windows': {'A': [15], 'B': [30], 'C': [15], 'D': [15, 30], 'E': [30, 15]},
   'lambdas': {'A': {(0,): 0.05, (1,): 0.2},
    'B': {(0,): 0.1, (1,): 0.3},
    'C': {(0,): 0.4, (1,): 0.2},
    'D': {(0, 0): 0.1, (0, 1): 0.3, (1, 0): 0.05, (1, 1): 0.2},
    'E': {(0, 0): 0.1, (0, 1): 0.02, (1, 0): 0.4, (1, 1): 0.1}}}.
    
\textbf{E} has the follow parameters:  
{'parents': {'A': ['A'],
    'B': ['A', 'C'],
    'C': ['C'],
    'D': ['A', 'E'],
    'E': ['C', 'D']},
   'windows': {'A': [15],
    'B': [30, 30],
    'C': [15],
    'D': [15, 30],
    'E': [15, 30]},
   'lambdas': {'A': {(0,): 0.1, (1,): 0.3},
    'B': {(0, 0): 0.01, (0, 1): 0.05, (1, 0): 0.1, (1, 1): 0.5},
    'C': {(0,): 0.2, (1,): 0.4},
    'D': {(0, 0): 0.05, (0, 1): 0.02, (1, 0): 0.2, (1, 1): 0.1},
    'E': {(0, 0): 0.1, (0, 1): 0.01, (1, 0): 0.3, (1, 1): 0.1}}}.

\begin{figure*} [htbp]
    \centering
    \includegraphics[width=0.8\textwidth]{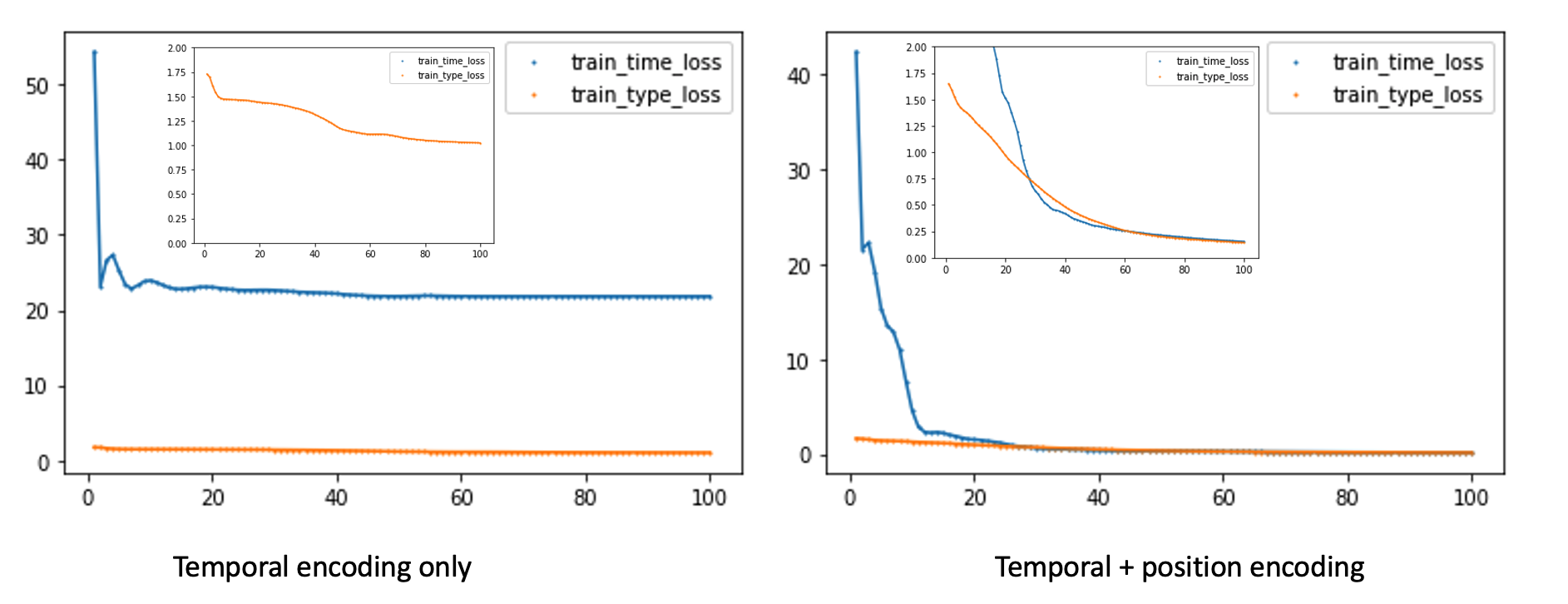}
    \caption{The effect of combined TE + PE in training.}
    \label{fig:pete}
\end{figure*}

\textbf{Combined Position Encoding (PE) and Temporal Encoding (TE).} We present a straightforward yet illustrative example demonstrating the effectiveness of the combined encoding scheme, which incorporates both position encoding and temporal encoding. In our experiment, we trained the model with only two samples, and the results clearly demonstrate improved overall training compared to using temporal encoding alone. Figure \ref{fig:pete} visually depicts the compromised optimization when employing only temporal encoding.

\section{Baseline Models and Implementation}
We provide a brief overview of the implementation of the baseline models.

T-RMTPP and T-ERPP: The original Recurrent Marked Temporal Point Process (RMTPP) and Event Recurrent Point Process (ERPP) models are based on recurrent neural networks (RNNs) or their variants. Our implementation builds upon a standard codebase \footnote{https://github.com/woshiyyya/ERPP-RMTPP}. However, to ensure a fair comparison, we replaced the RNN components with transformers. We trained these models using the recommended set of parameters in our experiments.

T-LNM: The original Lognormal Mixture (LNM) model, like the previous models, is RNN-based. To model the density of log (inter-event) times, we utilized 64 mixtures of lognormal components while keeping other parameters at their default setting \footnote{https://github.com/shchur/ifl-tpp}.

THP: The Transformer Hawkes Process (THP) models were implemented using attention-based transformers. We followed the recommended parameter settings for these models in our experiments.\footnote{https://github.com/SimiaoZuo/Transformer-Hawkes-Process}.

Initiator and NCE-MPP: The two model were implemented using continuous time LSTM. We followed the recommended parameter settings for these models in our experiments except we set 100 epochs to be consistent with other models. Initiator is implemented as NCE-binary module in the corresponding repo \footnote{https://github.com/hongyuanmei/nce-mpp}.

\section{Proof Sketch of Proposition 1: Transformer with Combined Temporal Encoding and Position Encoding}
We consider a general type of transformer architecture as described by Yun et al. (2019) in their work on transformers. Our proof demonstrates that a transformer with combined temporal encoding and position encoding can serve as a universal approximating function for any continuous function in the context of sequence-to-sequence tasks with compact support. This proof follows a similar structure to the proof of the transformer with position encoding (Theorem 3 in \cite{yun2019transformers}). 

Without loss of generality, let's consider a sequence with timestamps ${t_1, t_2, \ldots, t_n}$. We assume that $t_i$ is an integer value for each $i \in {1, 2, 3, \ldots, n}$, and that $t_i < t_j$ for all $i < j$. In case any $t_i$ values are in decimal form, we can multiply them by a constant factor without affecting the dynamics of the event sequence, thereby transforming each given timestamp to an integer.

We define a $d$-dimensional temporal encoding for the $n$ event epochs as follows:
\begin{equation*}
\mathbf{T} = 
 \begin{bmatrix}
0 & t_2-t_1 & \dots \ t_n -t_1 \\
0 & t_2-t_1 & \dots \ t_n -t_1 \\
\vdots & \vdots  & \vdots   &      \\
0 & t_2-t_1 & \dots \ t_n -t_1 \\
\end{bmatrix}   
\end{equation*}
Similarly, a $d$-dimensional position encoding for the $n$ event epochs is defined as:
\begin{equation*}
\mathbf{P} = 
 \begin{bmatrix}
0 & 1 & \dots \ n \\
0 & 1 & \dots \ n\\
\vdots & \vdots  & \vdots   &      \\
0 & 1 & \dots \ n \\
\end{bmatrix}   
\end{equation*}
The combined encoding is given by:
\begin{equation*}
\mathbf{PE + TE} = 
 \begin{bmatrix}
0 & t_2-t_1+1 & \dots \ t_n -t_1 +1\\
0 & t_2-t_1+1 & \dots \ t_n -t_1+1 \\
\vdots & \vdots  & \vdots   &      \\
0 & t_2-t_1+1 & \dots \ t_n -t_1 +1\\
\end{bmatrix}   
\end{equation*}

The strict temporal ordering of the timestamps guarantees that $t_i - t_1 + 1 < t_{i+1} - t_1 + 1$ for all $i \in \{1, 2, 3, \ldots, n-1\}$. This ensures that for all rows, the coordinates are monotonically increasing, and the input values can be partitioned into cubes.

The remaining steps of the proof closely follow those outlined in the proof of Theorem 3 in \cite{yun2019transformers}. This involves replacing $n$ with $t_n$ through quantization performed by feed-forward layers, contextual mapping achieved by attention layers, and function value mapping implemented by feed-forward layers.

\end{document}

\maketitle

\section{Synthetic Generators}
We generated datasets from Hawkes process and Proximal Graphical Event Model (PGEM). We describe the parameters used in our experiments to generate event datasets.

\paragraph{Hawkes-Exp.} We use a standard library to simulate 10-dimensional event sequences from the Hawkes dynamics with exponential decay\footnote{https://x-datainitiative.github.io/tick/}.The parameters which control the generation are baseline rate, decay coefficent, adjacency (infectivity matrix) and end time.
The four parameters for models A, B, C, D, E and F in our study are different from each other, and thus represent a variety of generating mechanisms which are implied by differing baseline rates, decaying patterns, triggering behaviors and sequence time horizons. The actual values for the 6 models used in our experiments are listed in the following for the sake of reproducibility. \\
\\
\textbf{A}. baseline = [0.1097627 , 0.14303787, 0.12055268, 0.10897664, 0.08473096,
       0.12917882, 0.08751744, 0.1783546 , 0.19273255, 0.0766883], decay = 2.5, infectivity = [[0.15037453, 0.10045448, 0.10789028, 0.17580114, 0.01349208,
        0.01654871, 0.00384014, 0.1581418 , 0.14779747, 0.16524382],
       [0.1858717 , 0.1517864 , 0.08765006, 0.14824807, 0.02246419,
        0.12154197, 0.02722749, 0.17942359, 0.0991161 , 0.07875789],
       [0.05024778, 0.14705235, 0.0866379 , 0.10796424, 0.0035688 ,
        0.11730922, 0.11625704, 0.11717598, 0.17924869, 0.12950002],
       [0.06828233, 0.08300669, 0.13250303, 0.01143879, 0.12664085,
        0.12737611, 0.03995854, 0.02448733, 0.05991018, 0.0690806 ],
       [0.10829905, 0.0833048 , 0.18772458, 0.01938165, 0.03967254,
        0.03063796, 0.12404668, 0.04810838, 0.0885677 , 0.04642443],
       [0.03019353, 0.02096386, 0.1246585 , 0.02624547, 0.03733743,
        0.07003299, 0.15593352, 0.01844271, 0.1591532 , 0.01825224],
       [0.18546165, 0.08901222, 0.18551894, 0.11487999, 0.14041038,
        0.00744305, 0.05371431, 0.02282927, 0.05624673, 0.02255028],
       [0.06039543, 0.07868212, 0.01218371, 0.13152315, 0.10761619,
        0.05040616, 0.09938195, 0.01784238, 0.10939112, 0.1765038 ],
       [0.06050668, 0.1267631 , 0.02503273, 0.13605401, 0.0549677 ,
        0.03479404, 0.11139803, 0.00381908, 0.15744288, 0.00089182],
       [0.12873957, 0.05128336, 0.13963744, 0.18275114, 0.04724637,
        0.10943116, 0.11244817, 0.10868939, 0.04237051, 0.18095826]] and end-time = 10. 
        
\textbf{B}. Baseline = [8.34044009e-02, 1.44064899e-01, 2.28749635e-05, 6.04665145e-02,
       2.93511782e-02, 1.84677190e-02, 3.72520423e-02, 6.91121454e-02,
       7.93534948e-02, 1.07763347e-01], decay = 2.5, 
      adjacency = [[0.        , 0.        , 0.03514877, 0.15096311, 0.        ,
        0.11526461, 0.07174169, 0.09604815, 0.02413487, 0.        ],
       [0.        , 0.        , 0.        , 0.        , 0.15066599,
        0.15379789, 0.01462053, 0.00671417, 0.        , 0.        ],
       [0.01690747, 0.07239546, 0.16467728, 0.        , 0.11894528,
        0.        , 0.11802102, 0.14348615, 0.00314406, 0.12896239],
       [0.17000181, 0.        , 0.        , 0.        , 0.        ,
        0.        , 0.        , 0.0504772 , 0.04947341, 0.        ],
       [0.        , 0.11670321, 0.03638242, 0.        , 0.        ,
        0.00917392, 0.        , 0.0252251 , 0.10131151, 0.1203002 ],
       [0.        , 0.07118317, 0.11937903, 0.07120436, 0.        ,
        0.        , 0.        , 0.08851807, 0.16239168, 0.10083865],
       [0.        , 0.02363421, 0.02394394, 0.1388041 , 0.06836732,
        0.02842716, 0.15945428, 0.        , 0.        , 0.12481123],
       [0.15185513, 0.        , 0.1290996 , 0.        , 0.        ,
        0.15401787, 0.        , 0.16587219, 0.11405672, 0.10687992],
       [0.        , 0.        , 0.07734744, 0.09943488, 0.07016556,
        0.04074891, 0.        , 0.        , 0.00049346, 0.10609756],
       [0.05615574, 0.09061013, 0.15230831, 0.06142066, 0.15619243,
        0.10716606, 0.00271994, 0.15978585, 0.11877677, 0.17145652]] and end-time = 40.

\textbf{C}. baseline = [0.08719898, 0.00518525, 0.1099325 , 0.08706448, 0.08407356,
       0.06606696, 0.04092973, 0.12385419, 0.05993093, 0.05336546], decay = 2.5, adjacency = [[0.        , 0.09623737, 0.        , 0.        , 0.        ,
        0.14283231, 0.        , 0.        , 0.        , 0.        ],
       [0.        , 0.        , 0.        , 0.        , 0.        ,
        0.        , 0.        , 0.        , 0.        , 0.        ],
       [0.        , 0.        , 0.        , 0.        , 0.        ,
        0.        , 0.14434229, 0.10548788, 0.        , 0.        ],
       [0.        , 0.        , 0.16178088, 0.        , 0.        ,
        0.        , 0.        , 0.        , 0.        , 0.        ],
       [0.        , 0.        , 0.        , 0.        , 0.        ,
        0.        , 0.        , 0.        , 0.        , 0.        ],
       [0.14554646, 0.        , 0.        , 0.        , 0.        ,
        0.09531432, 0.        , 0.        , 0.        , 0.        ],
       [0.        , 0.        , 0.        , 0.        , 0.        ,
        0.        , 0.        , 0.        , 0.        , 0.04899491],
       [0.        , 0.        , 0.        , 0.07048057, 0.        ,
        0.07546073, 0.        , 0.        , 0.        , 0.        ],
       [0.05697369, 0.        , 0.        , 0.        , 0.        ,
        0.        , 0.11709833, 0.        , 0.03100542, 0.        ],
       [0.        , 0.        , 0.        , 0.        , 0.        ,
        0.        , 0.04016475, 0.        , 0.10768499, 0.06297179]] and end-time = 80.
        
\textbf{D}. Baseline = [0.11015958, 0.14162956, 0.05818095, 0.10216552, 0.17858939,
       0.17925862, 0.02511706, 0.04144858, 0.01029344, 0.08816197], decay = [[8., 9., 2., 7., 3., 3., 2., 4., 6., 9.],
       [2., 9., 8., 9., 2., 1., 6., 5., 2., 6.],
       [5., 8., 7., 1., 1., 3., 5., 6., 9., 9.],
       [8., 6., 2., 2., 2., 6., 6., 8., 5., 4.],
       [1., 1., 1., 1., 3., 3., 8., 1., 6., 1.],
       [2., 5., 2., 3., 3., 5., 9., 1., 7., 1.],
       [5., 2., 6., 2., 9., 9., 8., 1., 1., 2.],
       [8., 9., 8., 5., 1., 1., 5., 4., 1., 9.],
       [3., 8., 3., 2., 4., 3., 5., 2., 3., 3.],
       [8., 4., 5., 2., 7., 8., 2., 1., 1., 6.]], adjacency = [[0.        , 0.14343731, 0.        , 0.11247478, 0.        ,
        0.        , 0.        , 0.09416725, 0.        , 0.        ],
       [0.        , 0.        , 0.        , 0.        , 0.        ,
        0.        , 0.        , 0.15805746, 0.        , 0.08262718],
       [0.        , 0.        , 0.03018515, 0.        , 0.        ,
        0.        , 0.        , 0.        , 0.        , 0.        ],
       [0.        , 0.11090719, 0.08158544, 0.        , 0.11109462,
        0.        , 0.        , 0.        , 0.        , 0.        ],
       [0.        , 0.        , 0.14264684, 0.        , 0.        ,
        0.11786607, 0.        , 0.        , 0.01101593, 0.        ],
       [0.04954495, 0.        , 0.        , 0.        , 0.        ,
        0.12385743, 0.        , 0.        , 0.02375575, 0.05345351],
       [0.        , 0.14941748, 0.02618691, 0.        , 0.13608937,
        0.        , 0.06263167, 0.        , 0.04097688, 0.14101171],
       [0.        , 0.11902986, 0.        , 0.04889382, 0.        ,
        0.        , 0.01569298, 0.03678315, 0.        , 0.        ],
       [0.05359555, 0.        , 0.        , 0.09188512, 0.        ,
        0.14255311, 0.        , 0.        , 0.        , 0.        ],
       [0.12792415, 0.05843994, 0.16156482, 0.11931973, 0.        ,
        0.00774966, 0.00947755, 0.        , 0.        , 0.        ]], and end-time = 50.

\textbf{E}. Baseline = [0.2, 0.2, 0.2, 0.2, 0.2, 0.2, 0.2, 0.2, 0.2, 0.2], decay = [[9., 4., 9., 9., 1., 6., 4., 6., 8., 7.],
       [1., 5., 8., 9., 2., 7., 3., 3., 2., 4.],
       [6., 9., 2., 9., 8., 9., 2., 1., 6., 5.],
       [2., 6., 5., 8., 7., 1., 1., 3., 5., 6.],
       [9., 9., 8., 6., 2., 2., 2., 6., 6., 8.],
       [5., 4., 1., 1., 1., 1., 3., 3., 8., 1.],
       [6., 1., 2., 5., 2., 3., 3., 5., 9., 1.],
       [7., 1., 5., 2., 6., 2., 9., 9., 8., 1.],
       [1., 2., 8., 9., 8., 5., 1., 1., 5., 4.],
       [1., 9., 3., 8., 3., 2., 4., 3., 5., 2.]],
      adjacency= [[0.02186539, 0.09356695, 0.        , 0.16101355, 0.11527002,
        0.09149395, 0.        , 0.        , 0.15672219, 0.        ],
       [0.        , 0.        , 0.14241135, 0.        , 0.11167029,
        0.        , 0.        , 0.        , 0.0934937 , 0.        ],
       [0.        , 0.        , 0.        , 0.        , 0.        ,
        0.        , 0.        , 0.        , 0.15692693, 0.        ],
       [0.08203618, 0.        , 0.        , 0.02996925, 0.        ,
        0.        , 0.        , 0.        , 0.        , 0.        ],
       [0.        , 0.        , 0.11011391, 0.08100189, 0.        ,
        0.11029999, 0.        , 0.        , 0.        , 0.        ],
       [0.        , 0.        , 0.        , 0.14162654, 0.        ,
        0.        , 0.11702301, 0.        , 0.        , 0.01093714],
       [0.        , 0.04919057, 0.        , 0.        , 0.        ,
        0.        , 0.12297152, 0.        , 0.        , 0.02358583],
       [0.05307117, 0.        , 0.14834875, 0.02599961, 0.        ,
        0.13511597, 0.        , 0.06218368, 0.        , 0.04068378],
       [0.1400031 , 0.        , 0.11817848, 0.        , 0.0485441 ,
        0.        , 0.        , 0.01558073, 0.03652006, 0.        ],
       [0.        , 0.05321219, 0.        , 0.        , 0.0912279 ,
        0.        , 0.14153347, 0.        , 0.        , 0.        ]] and end-time = 50. 
        
\textbf{F}. Baseline = [0.21736198, 0.11134775, 0.16980704, 0.33791045, 0.00188754,
       0.04862765, 0.26829963, 0.3303411 , 0.05468264, 0.23003733], decay = [[5., 1., 7., 3., 5., 2., 6., 4., 5., 5.],
       [4., 8., 2., 2., 8., 8., 1., 3., 4., 3.],
       [6., 9., 2., 1., 8., 7., 3., 1., 9., 3.],
       [6., 2., 9., 2., 6., 5., 3., 9., 4., 6.],
       [1., 4., 7., 4., 5., 8., 7., 4., 1., 5.],
       [5., 6., 8., 7., 7., 3., 5., 3., 8., 2.],
       [7., 7., 1., 8., 3., 4., 6., 5., 3., 5.],
       [4., 8., 1., 1., 6., 7., 7., 6., 7., 5.],
       [8., 4., 3., 4., 9., 8., 2., 6., 4., 1.],
       [7., 3., 4., 5., 9., 9., 6., 3., 8., 6.]], adjacency = [[0.15693854, 0.04896059, 0.0400508 , 0.        , 0.        ,
        0.13021228, 0.        , 0.10699903, 0.        , 0.15329807],
       [0.0784283 , 0.        , 0.        , 0.        , 0.        ,
        0.        , 0.00310706, 0.0090892 , 0.07758874, 0.        ],
       [0.01672489, 0.        , 0.        , 0.07851303, 0.        ,
        0.        , 0.12848331, 0.08859293, 0.        , 0.        ],
       [0.09984995, 0.        , 0.        , 0.10541925, 0.        ,
        0.08032527, 0.        , 0.        , 0.        , 0.        ],
       [0.14642469, 0.06629365, 0.        , 0.        , 0.        ,
        0.        , 0.11891738, 0.04166225, 0.09808829, 0.17638655],
       [0.00976324, 0.1100343 , 0.02003261, 0.        , 0.        ,
        0.05993539, 0.09739541, 0.        , 0.        , 0.        ],
       [0.        , 0.        , 0.        , 0.04672133, 0.16916   ,
        0.        , 0.17341419, 0.12078975, 0.14441602, 0.        ],
       [0.        , 0.17305542, 0.06927975, 0.        , 0.03408974,
        0.        , 0.08457162, 0.03787486, 0.10863292, 0.        ],
       [0.07186225, 0.05760593, 0.        , 0.        , 0.08042031,
        0.04403479, 0.1033595 , 0.17046747, 0.        , 0.05083523],
       [0.        , 0.10029222, 0.        , 0.1022067 , 0.        ,
        0.        , 0.0588527 , 0.        , 0.03530513, 0.        ]], and end-time = 40. 

\paragraph{PGEM.} We implement the PGEM generator by previous work \cite{bhattacharjya2018proximal} and generate 5-dimensional event datasets governed by the PGEM dynamics, which is signified by its piecewise conditional intensity function for each event type at a given time instance of a parental configuration observed at a particular window. The parameters are the conditional intensity (lambdas) for each event type given parental states, parental configuration (parents), windows for each parental state (windows) and end time which is 100 across all models. The parameters for models A, B, C, D and E  are listed in the following.  We point out to the readers that the 5 models share different graphs, which is implied by the parents for each event type (node).

\textbf{A} has the following parameters: {'parents': {'A': [],
    'B': [],
    'C': ['B'],
    'D': ['A', 'B'],
    'E': ['C']},
   'windows': {'A': [], 'B': [], 'C': [15], 'D': [15, 30], 'E': [15]},
   'lambdas': {'A': {(): 0.2},
    'B': {(): 0.05},
    'C': {(0,): 0.2, (1,): 0.3},
    'D': {(0, 0): 0.1, (0, 1): 0.05, (1, 0): 0.3, (1, 1): 0.2},
    'E': {(0,): 0.1, (1,): 0.3}}}

\textbf{B} has the follow parameters: {'parents': {'A': ['B'],
    'B': ['B'],
    'C': ['B'],
    'D': ['A'],
    'E': ['C']},
   'windows': {'A': [15], 'B': [30], 'C': [15], 'D': [30], 'E': [30]},
   'lambdas': {'A': {(0,): 0.3, (1,): 0.2},
    'B': {(0,): 0.2, (1,): 0.4},
    'C': {(0,): 0.4, (1,): 0.1},
    'D': {(0,): 0.05, (1,): 0.2},
    'E': {(0,): 0.1, (1,): 0.3}}}.

\textbf{C} has the follow parameters:
{'parents': {'A': ['B', 'D'],
    'B': [],
    'C': ['B', 'E'],
    'D': ['B'],
    'E': ['B']},
   'windows': {'A': [15, 30], 'B': [], 'C': [15, 30], 'D': [30], 'E': [30]},
   'lambdas': {'A': {(0, 0): 0.1, (0, 1): 0.05, (1, 0): 0.3, (1, 1): 0.2},
    'B': {(): 0.2},
    'C': {(0, 0): 0.2, (0, 1): 0.05, (1, 0): 0.4, (1, 1): 0.3},
    'D': {(0,): 0.1, (1,): 0.2},
    'E': {(0,): 0.1, (1,): 0.4}}}.
    
\textbf{D} has the follow parameters: 
{'parents': {'A': ['B'],
    'B': ['C'],
    'C': ['A'],
    'D': ['A', 'B'],
    'E': ['B', 'C']},
   'windows': {'A': [15], 'B': [30], 'C': [15], 'D': [15, 30], 'E': [30, 15]},
   'lambdas': {'A': {(0,): 0.05, (1,): 0.2},
    'B': {(0,): 0.1, (1,): 0.3},
    'C': {(0,): 0.4, (1,): 0.2},
    'D': {(0, 0): 0.1, (0, 1): 0.3, (1, 0): 0.05, (1, 1): 0.2},
    'E': {(0, 0): 0.1, (0, 1): 0.02, (1, 0): 0.4, (1, 1): 0.1}}}.
    
\textbf{E} has the follow parameters:  
{'parents': {'A': ['A'],
    'B': ['A', 'C'],
    'C': ['C'],
    'D': ['A', 'E'],
    'E': ['C', 'D']},
   'windows': {'A': [15],
    'B': [30, 30],
    'C': [15],
    'D': [15, 30],
    'E': [15, 30]},
   'lambdas': {'A': {(0,): 0.1, (1,): 0.3},
    'B': {(0, 0): 0.01, (0, 1): 0.05, (1, 0): 0.1, (1, 1): 0.5},
    'C': {(0,): 0.2, (1,): 0.4},
    'D': {(0, 0): 0.05, (0, 1): 0.02, (1, 0): 0.2, (1, 1): 0.1},
    'E': {(0, 0): 0.1, (0, 1): 0.01, (1, 0): 0.3, (1, 1): 0.1}}}.

\begin{figure*} [htbp]
    \centering
    \includegraphics[width=0.9\textwidth]{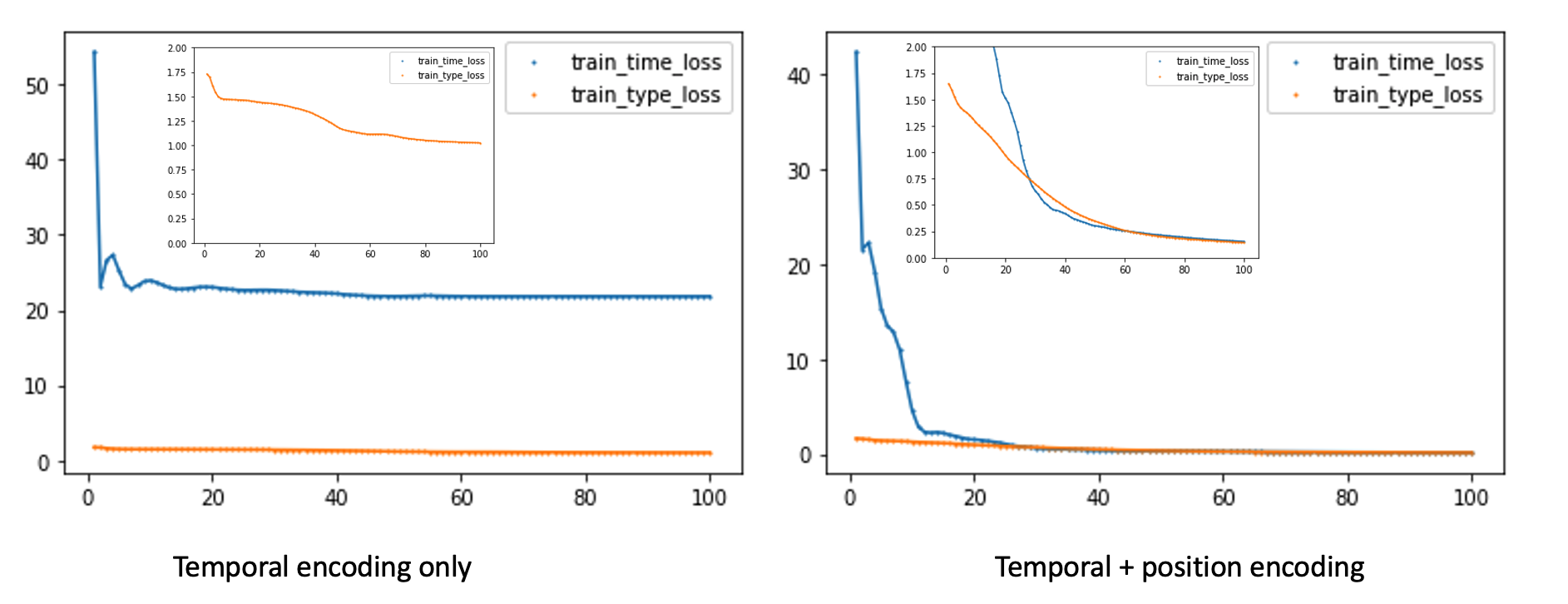}
    \caption{The effect of combined TE + PE in training.}
    \label{fig:pete}
\end{figure*}

\section{Model Implementation and (Pre-)Training}
% \textbf{Pretraining}. Our pre-training model adapts codes from \cite{zuo2020transformer} \footnote{https://github.com/SimiaoZuo/Transformer-Hawkes-Process/tree/master/transformer} and codes are included in the Supplement Material. %\footnote{Due to file size constraint, we only include hawkes pretraining dataset and 1 sample of target dataset ; we will post all datasets upon acceptance.}. 
% The procedure is fully described by Algorithm 1. We train our model via stochastic gradient descent and Adam optimizer \cite{kingma2014adam} is used for optimization. The default transformer architecture we employed are the following for pretraining: the number of blocks for multi-headed self-attention module is 4; the dimension of the value vector after attention has been applied is 512; the number attention heads is 4; the dimension of the hidden layer of the feed forward neural network 1024; the dimension of the value vector 512;  the dimension of the key vector 512; dropout is 0.1. We train 100 epochs with a learning rate of 0.0001. $\gamma$ is set to 1 for all experiments other than one on ACLED-india where we use 10. 
\textbf{Pretraining.} Our pretraining model is based on the code adaptation from \cite{zuo2020transformer}, which can be found in the Supplementary Material. The entire procedure is detailed in Algorithm 1 where a combined loss $\mathcal{L}_{event} = \mathcal{L}_{predict} + \lambda \mathcal{L}_{contr}$ is minimized for each batch where $\mathcal{L}_{predict}, \mathcal{L}_{contr}$ are from equation 3 and 4 respectively. We first randomly sample void event instances in-between every consecutive real events and then insert such void events. Then for each augment sequence, we randomly mask 15\% of the events and split into train and dev subsets of sequences. We train our model using stochastic gradient descent and employ the Adam optimizer for optimization. The default transformer architecture used for pretraining includes the following specifications: the multi-headed self-attention module consists of 4 blocks, the dimension of the value vector after attention is applied is 512, the number of attention heads is 4, the hidden layer of the feed-forward neural network has a dimension of 1024, the value vector has a dimension of 512, the key vector has a dimension of 512, and the dropout rate is set at 0.1. Our training process spans 100 epochs with a learning rate of 0.0001.
The outputs of the transformer model, denoted as $\mathbf{H}_i$'s, are linearly mapped to numerical values and probabilities in the regression and classification losses presented in equation 3. All our experiments are performed on a private server with TITAN RTX GPU. %More details can be found via https://idea.rpi.edu/IDEA\_Cluster\_Access .

\RestyleAlgo{ruled} \label{alg:training}
\begin{algorithm} 
\caption{Pretraining of Event-former}
\KwInput{Given dataset $S$ with $D$ sequences, each sequence $S_d$ with length $d_l$, $\{(t_i,y_i)\}_{i=1}^{d_l} $ , batch size $b$ } \\
$S' = []$ \\
\For{$d \gets1$ \KwTo $D$ }{
    seq' = [] \\
    \For{$i \gets1$ \KwTo $d_l-1$ }{ 
        $t'_i$ \sim Unif($t_i$,$t_{i+1}$) \\
        seq'.\textbf{append}(($t'_i$, null)) 
}
$seq_{new}$ = \textbf{merge}($S_d$, seq') \\
S'.\textbf{append}( $seq_{new}$ )} 
\For{$d \gets1$ \KwTo $D$ }{
    \For{$i \gets1$ \KwTo $d'_l-1$ }{ 
        mask \sim Bern(0.15) \\
        \If{$mask ==1$} {
        (0, null) \gets  (t'_i,y'_i) \\ }
}
}
\textbf{split}(S'_m) :=  $S'_{m,tr} = \{S'_{m,k}\}_{k=1}^{K}$, $S'_{m,dev} $ \\
\For{$epoch\gets1$ \KwTo $N$ }{
    \For{$iteration\gets1$ \KwTo $\lceil {\frac{K}{b}}  \rceil $  } { 
        $B'$ \sim $S'_{m,tr}$; Sample a batch of sequences  \\
        compute $\mathcal{L}_{event} (B')$ (Eq. 3) \\
        back-propagate with gradient $\nabla_{\theta,\phi} \mathcal{L}_{event}(B')$  ( $\theta$: Transformer model parameters, $\phi$: Weights and bias for regression and classification.) \\   
        update parameters of network $\theta,\phi $ \\
    }
    evaluate $L_{event}(S'_{m,dev})$, stop training if not improving in 5 epochs
    } \\
Return: Optimal parameters $\phi^{*}$, $\theta^{*}$
\end{algorithm}

\textbf{Fine-tuning.} For the fine-tuning process, we employ a feed-forward network consisting of three hidden layers, each with a dimension of 512. The representations of the training and development subsets are acquired from the pre-trained model and are held constant throughout the fine-tuning stage. We utilize the Adam optimizer with a learning rate chosen from the range of $\{0.001, 0.002\}$, along with a batch size of 32, for a total of 100 epochs. To prevent overfitting, we implement early stopping if there is no improvement in the loss $\mathcal{L}_{fine-tune}$ for 10 consecutive epochs. The learning rate is determined based on the model's performance on the development subset of the target data. %Furthermore, in all experiments, the value of $\alpha$ in equation 5 is fixed at 0.01.

\textbf{Hyperparameter Selection.} We discuss 4 important hyperparameters discussed in our paper. The three in pre-training are the trade-off between label and time prediction loss $\gamma$ in Equation 3, the temperature parameter $\omega$ in Equation 4, the trade-off parameter between prediction and contrastive loss $\lambda$; the fourth in fine tuning is the trade-off parameter between time and label in Equation 5. In all experiments, the hyper-parameters were selected from the best performing model on validation subset. Specifically, $\gamma$ was chosen from $\{0.01,0.1,1\}$ and empirically we found $\gamma=1$ worked well in our setting. $\omega$ could be considered as a fixed value at 1 since Equation 4 could be simplified to a more concise form by absorbing $\omega$ into $\lambda$ in practice. $lambda$ was given the range $\{0.1,0.01,0.001,0.0001\}$ and we found $0.01$ worked best. We followed early work \cite{zuo2020transformer} and fixed $\alpha$ at 0.01 in all fine tuning experiments. 

\textbf{Combined Position Encoding (PE) and Temporal Encoding (TE).} We present a straightforward yet illustrative example demonstrating the effectiveness of the combined encoding scheme, which incorporates both position encoding and temporal encoding. In our experiment, we trained the model with only two samples, and the results clearly demonstrate improved overall training compared to using temporal encoding alone. Figure \ref{fig:pete} visually depicts the compromised optimization when employing only temporal encoding.

\section{Baseline Models and Implementation}
In our study, we provide a brief overview of the implementation of the baseline models.

T-RMTPP and T-ERPP: The original Recurrent Marked Temporal Point Process (RMTPP) and Event Recurrent Point Process (ERPP) models are based on recurrent neural networks (RNNs) or their variants. Our implementation builds upon a standard codebase \footnote{https://github.com/woshiyyya/ERPP-RMTPP}. However, to ensure a fair comparison, we replaced the RNN components with transformers. We trained these models using the recommended set of parameters in our experiments.

T-LNM: The original Lognormal Mixture (LNM) model, like the previous models, is RNN-based. To model the density of log (inter-event) times, we utilized 64 mixtures of lognormal components while keeping other parameters at their default setting \footnote{https://github.com/shchur/ifl-tpp}.

THP: The Transformer Hawkes Process (THP) models were implemented using attention-based transformers. We followed the recommended parameter settings for these models in our experiments.\footnote{https://github.com/SimiaoZuo/Transformer-Hawkes-Process}.

Initiator and NCE-MPP: The two model were implemented using continuous time LSTM \cite{mei2016neural}. We followed the recommended parameter settings for these models in our experiments except we set 100 epochs to be consistent with other models. Initiator is implemented as NCE-binary module in the corresponding repo \footnote{https://github.com/hongyuanmei/nce-mpp/blob/main/ncempp/models/cont_time_cell.py}.

\section{Proof Sketch of Proposition 1: Transformer with Combined Temporal Encoding and Position Encoding}
We consider a general type of transformer architecture as described by Yun et al. (2019) in their work on transformers. Our proof demonstrates that a transformer with combined temporal encoding and position encoding can serve as a universal approximating function for any continuous function in the context of sequence-to-sequence tasks with compact support. This proof follows a similar structure to the proof of the transformer with position encoding (Theorem 3 in \cite{yun2019transformers}). 

Without loss of generality, let's consider a sequence with timestamps ${t_1, t_2, \ldots, t_n}$. We assume that $t_i$ is an integer value for each $i \in {1, 2, 3, \ldots, n}$, and that $t_i < t_j$ for all $i < j$. In case any $t_i$ values are in decimal form, we can multiply them by a constant factor without affecting the dynamics of the event sequence, thereby transforming each given timestamp to an integer.

We define a $d$-dimensional temporal encoding for the $n$ event epochs as follows:
\begin{equation*}
\mathbf{T} = 
 \begin{bmatrix}
0 & t_2-t_1 & \dots \ t_n -t_1 \\
0 & t_2-t_1 & \dots \ t_n -t_1 \\
\vdots & \vdots  & \vdots   &      \\
0 & t_2-t_1 & \dots \ t_n -t_1 \\
\end{bmatrix}   
\end{equation*}
Similarly, a $d$-dimensional position encoding for the $n$ event epochs is defined as:
\begin{equation*}
\mathbf{P} = 
 \begin{bmatrix}
0 & 1 & \dots \ n \\
0 & 1 & \dots \ n\\
\vdots & \vdots  & \vdots   &      \\
0 & 1 & \dots \ n \\
\end{bmatrix}   
\end{equation*}
The combined encoding is given by:
\begin{equation*}
\mathbf{PE + TE} = 
 \begin{bmatrix}
0 & t_2-t_1+1 & \dots \ t_n -t_1 +1\\
0 & t_2-t_1+1 & \dots \ t_n -t_1+1 \\
\vdots & \vdots  & \vdots   &      \\
0 & t_2-t_1+1 & \dots \ t_n -t_1 +1\\
\end{bmatrix}   
\end{equation*}

The strict temporal ordering of the timestamps guarantees that $t_i - t_1 + 1 < t_{i+1} - t_1 + 1$ for all $i \in \{1, 2, 3, \ldots, n-1\}$. This ensures that for all rows, the coordinates are monotonically increasing, and the input values can be partitioned into cubes.

The remaining steps of the proof closely follow those outlined in the proof of Theorem 3 in \cite{yun2019transformers}. This involves replacing $n$ with $t_n$ through quantization performed by feed-forward layers, contextual mapping achieved by attention layers, and function value mapping implemented by feed-forward layers.

% The strict temporal order of timestamps $t_i-t_1 + 1 < t_{i+1} -t_1 + 1 $for all $i \in {1,2,3,...,n-1}$. This guarantees for all rows the coordinates are monotonically increasing and input values can be partitioned into cubes. The rest of the proof follows directly from the proof of Theorem 3 in \cite{yun2019transformers} by replacing $n$ with $t_n$ by performing quantization by feed-forward layers, contextual mapping by attention layers and function value mapping by feed-forward layers. 

\section{Clarification, Limitations And Societal Impact}

\paragraph{Additional rationale for pre-training and fine-tuning.} In the context of dynamic event datasets, the role of labels extends beyond being mere targets for prediction, as observed in static image and text classification tasks. Here, labels in dynamic event datasets also serve as dynamic "features" in addition to their target role. Self-supervised representation learning emerges as a particularly effective approach for achieving a concise representation of the dynamic event history, utilizing event labels and timestamps as raw features. Our objective is to capitalize on a large volume of event sequences to acquire high-quality representations of event dynamics, especially for downstream tasks with limited data availability. Consequently, learning the representation of historical context and transferring a sequence of such representations that encapsulate event dynamics differs from learning representations for tabular, image, or text data. A similar motivation can be observed in time series transfer learning \cite{zerveas2021transformer}, where self-supervised techniques demonstrate their effectiveness in acquiring compact dynamic features from raw historical values in a time series.

\paragraph{Advantages of homogeneous transfer.} The homogeneous transfer problem, where pre-training on open domains and target task domains share the same set of event types, is not an oversimplified scenario. In fact, a significant portion of transfer learning research focuses on this type of transfer \cite{9134370}. In our study, we also consider this scenario, where the event types are shared across pre-training and fine-tuning; however, the datasets themselves may differ, resulting in varying underlying dynamics between datasets. We have presented several practical applications in our paper, supported by experimental evidence and thorough exposition. In the case of alignment of event types, we utilize shared embedding matrices for event types during both pre-training and fine-tuning in our homogeneous transfer setting. Therefore, the challenge of aligning event types does not arise. However, it is worth noting that addressing the alignment issue becomes relevant for heterogeneous transfer, which is an avenue we are currently exploring.
 
\paragraph{Impact of self-supervised pretraining for TPP.} Our self-supervised pretraining approach for temporal point process (TPP) models has certain considerations and potential societal implications that warrant attention. Firstly, the availability and representativeness of large-scale event streams, upon which our pretraining relies, may vary, posing challenges in accessing diverse real-world scenarios. Additionally, self-supervised training can be computationally demanding, necessitating substantial computational resources and time investments. Moreover, we acknowledge the potential risks associated with biases present in some data streams, which could lead to biased model predictions and the potential reinforcement of societal inequalities. Therefore, caution should be exercised when utilizing our model, and careful data curation and preprocessing are essential to mitigate such biases. Furthermore, the widespread adoption of self-supervised training for TPP models has implications for domains like healthcare and finance. To prevent potential negative consequences and societal harm, ethical considerations encompassing privacy protection, fairness, and transparency must be addressed. While self-supervised training holds promising opportunities, its limitations and societal impact demand meticulous consideration to ensure responsible harnessing of its potential and its beneficial application in real-world contexts.

% Our method focuses on potentially long event sequences with
% multiple occurrences of all events. We assume proximal historical influence, along with strong
% ignorability and no confounding. The usage of causal models should be cautioned, to draw potentially
% harmful conclusions. In particular, the window sizes in our formulation, which can be subject to
% human inputs, can induce biases and provide different conclusions with different values.

%% The file named.bst is a bibliography style file for BibTeX 0.99c
\bibliography{neurip23}
\bibliographystyle{unsrt}

%% file: math_commands.tex
\usepackage{amsmath,amsfonts,bm}

\def\eqref#1{equation~\ref{#1}}
\def\1{\bm{1}}

\DeclareMathAlphabet{\mathsfit}{\encodingdefault}{\sfdefault}{m}{sl}
\SetMathAlphabet{\mathsfit}{bold}{\encodingdefault}{\sfdefault}{bx}{n}

\newcommand{\E}{\mathbb{E}}

%% file: main.bbl
%%% -*-BibTeX-*-
%%% Do NOT edit. File created by BibTeX with style
%%% ACM-Reference-Format-Journals [18-Jan-2012].

\begin{thebibliography}{34}

%%% ====================================================================
%%% NOTE TO THE USER: you can override these defaults by providing
%%% customized versions of any of these macros before the \bibliography
%%% command.  Each of them MUST provide its own final punctuation,
%%% except for \shownote{}, \showDOI{}, and \showURL{}.  The latter two
%%% do not use final punctuation, in order to avoid confusing it with
%%% the Web address.
%%%
%%% To suppress output of a particular field, define its macro to expand
%%% to an empty string, or better, \unskip, like this:
%%%
%%% \newcommand{\showDOI}[1]{\unskip}   % LaTeX syntax
%%%
%%% \def \showDOI #1{\unskip}           % plain TeX syntax
%%%
%%% ====================================================================

\ifx \showCODEN    \undefined \def \showCODEN     #1{\unskip}     \fi
\ifx \showDOI      \undefined \def \showDOI       #1{#1}\fi
\ifx \showISBNx    \undefined \def \showISBNx     #1{\unskip}     \fi
\ifx \showISBNxiii \undefined \def \showISBNxiii  #1{\unskip}     \fi
\ifx \showISSN     \undefined \def \showISSN      #1{\unskip}     \fi
\ifx \showLCCN     \undefined \def \showLCCN      #1{\unskip}     \fi
\ifx \shownote     \undefined \def \shownote      #1{#1}          \fi
\ifx \showarticletitle \undefined \def \showarticletitle #1{#1}   \fi
\ifx \showURL      \undefined \def \showURL       {\relax}        \fi
% The following commands are used for tagged output and should be
% invisible to TeX
\providecommand\bibfield[2]{#2}
\providecommand\bibinfo[2]{#2}
\providecommand\natexlab[1]{#1}
\providecommand\showeprint[2][]{arXiv:#2}

\bibitem[Bacry et~al\mbox{.}(2015)]%
        {bacry2015hawkes}
\bibfield{author}{\bibinfo{person}{Emmanuel Bacry}, \bibinfo{person}{Iacopo Mastromatteo}, {and} \bibinfo{person}{Jean-Fran{\c{c}}ois Muzy}.} \bibinfo{year}{2015}\natexlab{}.
\newblock \showarticletitle{Hawkes processes in finance}.
\newblock \bibinfo{journal}{\emph{Market Microstructure and Liquidity}} \bibinfo{volume}{1}, \bibinfo{number}{01} (\bibinfo{year}{2015}), \bibinfo{pages}{1550005}.
\newblock


\bibitem[Bhattacharjya et~al\mbox{.}(2018)]%
        {bhattacharjya2018proximal}
\bibfield{author}{\bibinfo{person}{Debarun Bhattacharjya}, \bibinfo{person}{Dharmashankar Subramanian}, {and} \bibinfo{person}{Tian Gao}.} \bibinfo{year}{2018}\natexlab{}.
\newblock \showarticletitle{Proximal graphical event models}.
\newblock \bibinfo{journal}{\emph{Advances in Neural Information Processing Systems}}  \bibinfo{volume}{31} (\bibinfo{year}{2018}), \bibinfo{pages}{8147--8156}.
\newblock


\bibitem[Bommasani et~al\mbox{.}(2021)]%
        {bommasani}
\bibfield{author}{\bibinfo{person}{Rishi Bommasani}, \bibinfo{person}{Drew~A. Hudson}, \bibinfo{person}{Ehsan Adeli}, \bibinfo{person}{Russ Altman}, \bibinfo{person}{Simran Arora}, \bibinfo{person}{Sydney von Arx}, \bibinfo{person}{Michael~S. Bernstein}, \bibinfo{person}{Jeannette Bohg}, \bibinfo{person}{Antoine Bosselut}, \bibinfo{person}{Emma Brunskill}, \bibinfo{person}{Erik Brynjolfsson}, \bibinfo{person}{Shyamal Buch}, \bibinfo{person}{Dallas Card}, \bibinfo{person}{Rodrigo Castellon}, \bibinfo{person}{Niladri Chatterji}, \bibinfo{person}{Annie Chen}, \bibinfo{person}{Kathleen Creel}, \bibinfo{person}{Jared~Quincy Davis}, \bibinfo{person}{Dora Demszky}, \bibinfo{person}{Chris Donahue}, \bibinfo{person}{Moussa Doumbouya}, \bibinfo{person}{Esin Durmus}, \bibinfo{person}{Stefano Ermon}, \bibinfo{person}{John Etchemendy}, \bibinfo{person}{Kawin Ethayarajh}, \bibinfo{person}{Li Fei-Fei}, \bibinfo{person}{Chelsea Finn}, \bibinfo{person}{Trevor Gale}, \bibinfo{person}{Lauren Gillespie}, \bibinfo{person}{Karan
  Goel}, \bibinfo{person}{Noah Goodman}, \bibinfo{person}{Shelby Grossman}, \bibinfo{person}{Neel Guha}, \bibinfo{person}{Tatsunori Hashimoto}, \bibinfo{person}{Peter Henderson}, \bibinfo{person}{John Hewitt}, \bibinfo{person}{Daniel~E. Ho}, \bibinfo{person}{Jenny Hong}, \bibinfo{person}{Kyle Hsu}, \bibinfo{person}{Jing Huang}, \bibinfo{person}{Thomas Icard}, \bibinfo{person}{Saahil Jain}, \bibinfo{person}{Dan Jurafsky}, \bibinfo{person}{Pratyusha Kalluri}, \bibinfo{person}{Siddharth Karamcheti}, \bibinfo{person}{Geoff Keeling}, \bibinfo{person}{Fereshte Khani}, \bibinfo{person}{Omar Khattab}, \bibinfo{person}{Pang~Wei Koh}, \bibinfo{person}{Mark Krass}, \bibinfo{person}{Ranjay Krishna}, \bibinfo{person}{Rohith Kuditipudi}, \bibinfo{person}{Ananya Kumar}, \bibinfo{person}{Faisal Ladhak}, \bibinfo{person}{Mina Lee}, \bibinfo{person}{Tony Lee}, \bibinfo{person}{Jure Leskovec}, \bibinfo{person}{Isabelle Levent}, \bibinfo{person}{Xiang~Lisa Li}, \bibinfo{person}{Xuechen Li}, \bibinfo{person}{Tengyu Ma},
  \bibinfo{person}{Ali Malik}, \bibinfo{person}{Christopher~D. Manning}, \bibinfo{person}{Suvir Mirchandani}, \bibinfo{person}{Eric Mitchell}, \bibinfo{person}{Zanele Munyikwa}, \bibinfo{person}{Suraj Nair}, \bibinfo{person}{Avanika Narayan}, \bibinfo{person}{Deepak Narayanan}, \bibinfo{person}{Ben Newman}, \bibinfo{person}{Allen Nie}, \bibinfo{person}{Juan~Carlos Niebles}, \bibinfo{person}{Hamed Nilforoshan}, \bibinfo{person}{Julian Nyarko}, \bibinfo{person}{Giray Ogut}, \bibinfo{person}{Laurel Orr}, \bibinfo{person}{Isabel Papadimitriou}, \bibinfo{person}{Joon~Sung Park}, \bibinfo{person}{Chris Piech}, \bibinfo{person}{Eva Portelance}, \bibinfo{person}{Christopher Potts}, \bibinfo{person}{Aditi Raghunathan}, \bibinfo{person}{Rob Reich}, \bibinfo{person}{Hongyu Ren}, \bibinfo{person}{Frieda Rong}, \bibinfo{person}{Yusuf Roohani}, \bibinfo{person}{Camilo Ruiz}, \bibinfo{person}{Jack Ryan}, \bibinfo{person}{Christopher Ré}, \bibinfo{person}{Dorsa Sadigh}, \bibinfo{person}{Shiori Sagawa},
  \bibinfo{person}{Keshav Santhanam}, \bibinfo{person}{Andy Shih}, \bibinfo{person}{Krishnan Srinivasan}, \bibinfo{person}{Alex Tamkin}, \bibinfo{person}{Rohan Taori}, \bibinfo{person}{Armin~W. Thomas}, \bibinfo{person}{Florian Tramèr}, \bibinfo{person}{Rose~E. Wang}, \bibinfo{person}{William Wang}, \bibinfo{person}{Bohan Wu}, \bibinfo{person}{Jiajun Wu}, \bibinfo{person}{Yuhuai Wu}, \bibinfo{person}{Sang~Michael Xie}, \bibinfo{person}{Michihiro Yasunaga}, \bibinfo{person}{Jiaxuan You}, \bibinfo{person}{Matei Zaharia}, \bibinfo{person}{Michael Zhang}, \bibinfo{person}{Tianyi Zhang}, \bibinfo{person}{Xikun Zhang}, \bibinfo{person}{Yuhui Zhang}, \bibinfo{person}{Lucia Zheng}, \bibinfo{person}{Kaitlyn Zhou}, {and} \bibinfo{person}{Percy Liang}.} \bibinfo{year}{2021}\natexlab{}.
\newblock \showarticletitle{On the Opportunities and Risks of Foundation Models}.
\newblock \bibinfo{journal}{\emph{arXiv preprint arXiv:2108.07258}} (\bibinfo{year}{2021}).
\newblock


\bibitem[Brown et~al\mbox{.}(2020)]%
        {brown2020language}
\bibfield{author}{\bibinfo{person}{Tom Brown}, \bibinfo{person}{Benjamin Mann}, \bibinfo{person}{Nick Ryder}, \bibinfo{person}{Melanie Subbiah}, \bibinfo{person}{Jared~D Kaplan}, \bibinfo{person}{Prafulla Dhariwal}, \bibinfo{person}{Arvind Neelakantan}, \bibinfo{person}{Pranav Shyam}, \bibinfo{person}{Girish Sastry}, \bibinfo{person}{Amanda Askell}, {et~al\mbox{.}}} \bibinfo{year}{2020}\natexlab{}.
\newblock \showarticletitle{Language models are few-shot learners}.
\newblock \bibinfo{journal}{\emph{Advances in Neural Information Processing Systems}}  \bibinfo{volume}{33} (\bibinfo{year}{2020}), \bibinfo{pages}{1877--1901}.
\newblock


\bibitem[Daley and Jones(2003)]%
        {daley2003introduction}
\bibfield{author}{\bibinfo{person}{Daryl~J Daley} {and} \bibinfo{person}{D~Vere Jones}.} \bibinfo{year}{2003}\natexlab{}.
\newblock \bibinfo{booktitle}{\emph{An Introduction to the Theory of Point Processes: Elementary Theory of Point Processes}}.
\newblock \bibinfo{publisher}{Springer}.
\newblock


\bibitem[Devlin et~al\mbox{.}(2018)]%
        {devlin2018bert}
\bibfield{author}{\bibinfo{person}{Jacob Devlin}, \bibinfo{person}{Ming-Wei Chang}, \bibinfo{person}{Kenton Lee}, {and} \bibinfo{person}{Kristina Toutanova}.} \bibinfo{year}{2018}\natexlab{}.
\newblock \showarticletitle{{BERT}: Pre-training of deep bidirectional transformers for language understanding}.
\newblock \bibinfo{journal}{\emph{arXiv preprint arXiv:1810.04805}} (\bibinfo{year}{2018}).
\newblock


\bibitem[Du et~al\mbox{.}(2016)]%
        {du2016recurrent}
\bibfield{author}{\bibinfo{person}{Nan Du}, \bibinfo{person}{Hanjun Dai}, \bibinfo{person}{Rakshit Trivedi}, \bibinfo{person}{Utkarsh Upadhyay}, \bibinfo{person}{Manuel Gomez-Rodriguez}, {and} \bibinfo{person}{Le Song}.} \bibinfo{year}{2016}\natexlab{}.
\newblock \showarticletitle{Recurrent marked temporal point processes: Embedding event history to vector}. In \bibinfo{booktitle}{\emph{Proceedings of the 22nd ACM SIGKDD International Conference on Knowledge Discovery and Data Mining}}. \bibinfo{pages}{1555--1564}.
\newblock


\bibitem[Floridi and Chiriatti(2020)]%
        {floridi2020gpt}
\bibfield{author}{\bibinfo{person}{Luciano Floridi} {and} \bibinfo{person}{Massimo Chiriatti}.} \bibinfo{year}{2020}\natexlab{}.
\newblock \showarticletitle{{GPT-3}: Its nature, scope, limits, and consequences}.
\newblock \bibinfo{journal}{\emph{Minds and Machines}} \bibinfo{volume}{30}, \bibinfo{number}{4} (\bibinfo{year}{2020}), \bibinfo{pages}{681--694}.
\newblock


\bibitem[Gao et~al\mbox{.}(2020)]%
        {gao2020multi}
\bibfield{author}{\bibinfo{person}{Tian Gao}, \bibinfo{person}{Dharmashankar Subramanian}, \bibinfo{person}{Karthikeyan Shanmugam}, \bibinfo{person}{Debarun Bhattacharjya}, {and} \bibinfo{person}{Nicholas Mattei}.} \bibinfo{year}{2020}\natexlab{}.
\newblock \showarticletitle{A multi-channel neural graphical event model with negative evidence}. In \bibinfo{booktitle}{\emph{Proceedings of the AAAI Conference on Artificial Intelligence}}, Vol.~\bibinfo{volume}{34}. \bibinfo{pages}{3946--3953}.
\newblock


\bibitem[Gu(2021)]%
        {gu2021attentive}
\bibfield{author}{\bibinfo{person}{Yulong Gu}.} \bibinfo{year}{2021}\natexlab{}.
\newblock \showarticletitle{Attentive Neural Point Processes for Event Forecasting}. In \bibinfo{booktitle}{\emph{Proceedings of the AAAI Conference on Artificial Intelligence}}, Vol.~\bibinfo{volume}{35}. \bibinfo{pages}{7592--7600}.
\newblock


\bibitem[Gunawardana and Meek(2016)]%
        {gunawardana2016universal}
\bibfield{author}{\bibinfo{person}{Asela Gunawardana} {and} \bibinfo{person}{Chris Meek}.} \bibinfo{year}{2016}\natexlab{}.
\newblock \showarticletitle{Universal models of multivariate temporal point processes}. In \bibinfo{booktitle}{\emph{Artificial Intelligence and Statistics}}. PMLR, \bibinfo{pages}{556--563}.
\newblock


\bibitem[Guo et~al\mbox{.}(2018)]%
        {guo2018initiator}
\bibfield{author}{\bibinfo{person}{Ruocheng Guo}, \bibinfo{person}{Jundong Li}, {and} \bibinfo{person}{Huan Liu}.} \bibinfo{year}{2018}\natexlab{}.
\newblock \showarticletitle{INITIATOR: Noise-contrastive estimation for marked temporal point process}. In \bibinfo{booktitle}{\emph{Proceedings of the International Joint Conference on Artificial Intelligence}}. \bibinfo{pages}{2191--2197}.
\newblock


\bibitem[Hawkes(1971)]%
        {hawkes1971spectra}
\bibfield{author}{\bibinfo{person}{Alan~G Hawkes}.} \bibinfo{year}{1971}\natexlab{}.
\newblock \showarticletitle{Spectra of Some Self-exciting and Mutually Exciting Point Processes}.
\newblock \bibinfo{journal}{\emph{Biometrika}} \bibinfo{volume}{58}, \bibinfo{number}{1} (\bibinfo{year}{1971}), \bibinfo{pages}{83--90}.
\newblock


\bibitem[Jaiswal et~al\mbox{.}(2020)]%
        {jaiswal2020survey}
\bibfield{author}{\bibinfo{person}{Ashish Jaiswal}, \bibinfo{person}{Ashwin~Ramesh Babu}, \bibinfo{person}{Mohammad~Zaki Zadeh}, \bibinfo{person}{Debapriya Banerjee}, {and} \bibinfo{person}{Fillia Makedon}.} \bibinfo{year}{2020}\natexlab{}.
\newblock \showarticletitle{A survey on contrastive self-supervised learning}.
\newblock \bibinfo{journal}{\emph{Technologies}} \bibinfo{volume}{9}, \bibinfo{number}{1} (\bibinfo{year}{2020}), \bibinfo{pages}{2}.
\newblock


\bibitem[Kim et~al\mbox{.}(2017)]%
        {kim2017read}
\bibfield{author}{\bibinfo{person}{Hideaki Kim}, \bibinfo{person}{Tomoharu Iwata}, \bibinfo{person}{Yasuhiro Fujiwara}, {and} \bibinfo{person}{Naonori Ueda}.} \bibinfo{year}{2017}\natexlab{}.
\newblock \showarticletitle{Read the silence: Well-timed recommendation via admixture marked point processes}. In \bibinfo{booktitle}{\emph{Proceedings of the AAAI Conference on Artificial Intelligence}}, Vol.~\bibinfo{volume}{31}.
\newblock


\bibitem[Kingma and Ba(2014)]%
        {kingma2014adam}
\bibfield{author}{\bibinfo{person}{Diederik~P Kingma} {and} \bibinfo{person}{Jimmy Ba}.} \bibinfo{year}{2014}\natexlab{}.
\newblock \showarticletitle{Adam: A method for stochastic optimization}.
\newblock \bibinfo{journal}{\emph{arXiv preprint arXiv:1412.6980}} (\bibinfo{year}{2014}).
\newblock


\bibitem[Lian et~al\mbox{.}(2015)]%
        {lian2015multitask}
\bibfield{author}{\bibinfo{person}{Wenzhao Lian}, \bibinfo{person}{Ricardo Henao}, \bibinfo{person}{Vinayak Rao}, \bibinfo{person}{Joseph Lucas}, {and} \bibinfo{person}{Lawrence Carin}.} \bibinfo{year}{2015}\natexlab{}.
\newblock \showarticletitle{A multitask point process predictive model}. In \bibinfo{booktitle}{\emph{International Conference on Machine Learning}}. PMLR, \bibinfo{pages}{2030--2038}.
\newblock


\bibitem[Mei and Eisner(2016)]%
        {mei2016neural}
\bibfield{author}{\bibinfo{person}{Hongyuan Mei} {and} \bibinfo{person}{Jason Eisner}.} \bibinfo{year}{2016}\natexlab{}.
\newblock \showarticletitle{The Neural {H}awkes Process: A Neurally Self-modulating Multivariate Point Process}.
\newblock \bibinfo{journal}{\emph{arXiv preprint arXiv:1612.09328}} (\bibinfo{year}{2016}).
\newblock


\bibitem[Mei et~al\mbox{.}(2020)]%
        {mei2020noise}
\bibfield{author}{\bibinfo{person}{Hongyuan Mei}, \bibinfo{person}{Tom Wan}, {and} \bibinfo{person}{Jason Eisner}.} \bibinfo{year}{2020}\natexlab{}.
\newblock \showarticletitle{Noise-contrastive estimation for multivariate point processes}.
\newblock \bibinfo{journal}{\emph{Advances in Neural Information Processing Systems}}  \bibinfo{volume}{33} (\bibinfo{year}{2020}), \bibinfo{pages}{5204--5214}.
\newblock


\bibitem[Mei et~al\mbox{.}(2022)]%
        {mei2022transformer}
\bibfield{author}{\bibinfo{person}{Hongyuan Mei}, \bibinfo{person}{Chenghao Yang}, {and} \bibinfo{person}{Jason Eisner}.} \bibinfo{year}{2022}\natexlab{}.
\newblock \showarticletitle{Transformer embeddings of irregularly spaced events and their participants}. In \bibinfo{booktitle}{\emph{International Conference on Learning Representations}}.
\newblock


\bibitem[Monvoisin and Leray(2019)]%
        {monvoisin2019multi}
\bibfield{author}{\bibinfo{person}{Mathilde Monvoisin} {and} \bibinfo{person}{Philippe Leray}.} \bibinfo{year}{2019}\natexlab{}.
\newblock \showarticletitle{Multi-task transfer learning for timescale graphical event models}. In \bibinfo{booktitle}{\emph{European Conference on Symbolic and Quantitative Approaches with Uncertainty}}. Springer, \bibinfo{pages}{313--323}.
\newblock


\bibitem[Omi et~al\mbox{.}(2019)]%
        {omi2019fully}
\bibfield{author}{\bibinfo{person}{Takahiro Omi}, \bibinfo{person}{Naonori Ueda}, {and} \bibinfo{person}{Kazuyuki Aihara}.} \bibinfo{year}{2019}\natexlab{}.
\newblock \showarticletitle{Fully Neural Network Based Model for General Temporal Point Processes}.
\newblock \bibinfo{journal}{\emph{arXiv preprint arXiv:1905.09690}} (\bibinfo{year}{2019}).
\newblock


\bibitem[Shchur et~al\mbox{.}(2019)]%
        {shchur2019intensity}
\bibfield{author}{\bibinfo{person}{Oleksandr Shchur}, \bibinfo{person}{Marin Bilo{\v{s}}}, {and} \bibinfo{person}{Stephan G{\"u}nnemann}.} \bibinfo{year}{2019}\natexlab{}.
\newblock \showarticletitle{Intensity-free Learning of Temporal Point Processes}.
\newblock \bibinfo{journal}{\emph{arXiv preprint arXiv:1909.12127}} (\bibinfo{year}{2019}).
\newblock


\bibitem[Vaswani et~al\mbox{.}(2017)]%
        {vaswani2017attention}
\bibfield{author}{\bibinfo{person}{Ashish Vaswani}, \bibinfo{person}{Noam Shazeer}, \bibinfo{person}{Niki Parmar}, \bibinfo{person}{Jakob Uszkoreit}, \bibinfo{person}{Llion Jones}, \bibinfo{person}{Aidan~N Gomez}, \bibinfo{person}{{\L}ukasz Kaiser}, {and} \bibinfo{person}{Illia Polosukhin}.} \bibinfo{year}{2017}\natexlab{}.
\newblock \showarticletitle{Attention is all you need}.
\newblock \bibinfo{journal}{\emph{Advances in Neural Information Processing Systems}}  \bibinfo{volume}{30} (\bibinfo{year}{2017}).
\newblock


\bibitem[Wong and Chung(2019)]%
        {wong2019visualizing}
\bibfield{author}{\bibinfo{person}{Kwan~Yeung Wong} {and} \bibinfo{person}{Fu-lai Chung}.} \bibinfo{year}{2019}\natexlab{}.
\newblock \showarticletitle{Visualizing time series data with temporal matching based t-{SNE}}. In \bibinfo{booktitle}{\emph{Proceedings of the International Joint Conference on Neural Networks}}. IEEE, \bibinfo{pages}{1--8}.
\newblock


\bibitem[Xiao et~al\mbox{.}(2019)]%
        {xiao2019learning}
\bibfield{author}{\bibinfo{person}{Shuai Xiao}, \bibinfo{person}{Junchi Yan}, \bibinfo{person}{Mehrdad Farajtabar}, \bibinfo{person}{Le Song}, \bibinfo{person}{Xiaokang Yang}, {and} \bibinfo{person}{Hongyuan Zha}.} \bibinfo{year}{2019}\natexlab{}.
\newblock \showarticletitle{Learning time series associated event sequences with recurrent point process networks}.
\newblock \bibinfo{journal}{\emph{IEEE Transactions on Neural Networks and Learning Systems}} \bibinfo{volume}{30}, \bibinfo{number}{10} (\bibinfo{year}{2019}), \bibinfo{pages}{3124--3136}.
\newblock


\bibitem[Xiao et~al\mbox{.}(2017)]%
        {xiao2017modeling}
\bibfield{author}{\bibinfo{person}{Shuai Xiao}, \bibinfo{person}{Junchi Yan}, \bibinfo{person}{Xiaokang Yang}, \bibinfo{person}{Hongyuan Zha}, {and} \bibinfo{person}{Stephen~M Chu}.} \bibinfo{year}{2017}\natexlab{}.
\newblock \showarticletitle{Modeling the Intensity Function of Point Process Via Recurrent Neural Networks}. In \bibinfo{booktitle}{\emph{Proceedings of the Conference on Artificial Intelligence (AAAI)}}. \bibinfo{pages}{1597--1603}.
\newblock


\bibitem[Yang et~al\mbox{.}(2021)]%
        {yang2021voice2series}
\bibfield{author}{\bibinfo{person}{Chao-Han~Huck Yang}, \bibinfo{person}{Yun-Yun Tsai}, {and} \bibinfo{person}{Pin-Yu Chen}.} \bibinfo{year}{2021}\natexlab{}.
\newblock \showarticletitle{Voice2series: Reprogramming acoustic models for time series classification}. In \bibinfo{booktitle}{\emph{International Conference on Machine Learning}}. PMLR, \bibinfo{pages}{11808--11819}.
\newblock


\bibitem[Yun et~al\mbox{.}(2019)]%
        {yun2019transformers}
\bibfield{author}{\bibinfo{person}{Chulhee Yun}, \bibinfo{person}{Srinadh Bhojanapalli}, \bibinfo{person}{Ankit~Singh Rawat}, \bibinfo{person}{Sashank~J Reddi}, {and} \bibinfo{person}{Sanjiv Kumar}.} \bibinfo{year}{2019}\natexlab{}.
\newblock \showarticletitle{Are transformers universal approximators of sequence-to-sequence functions?}
\newblock \bibinfo{journal}{\emph{arXiv preprint arXiv:1912.10077}} (\bibinfo{year}{2019}).
\newblock


\bibitem[Zerveas et~al\mbox{.}(2021)]%
        {zerveas2021transformer}
\bibfield{author}{\bibinfo{person}{George Zerveas}, \bibinfo{person}{Srideepika Jayaraman}, \bibinfo{person}{Dhaval Patel}, \bibinfo{person}{Anuradha Bhamidipaty}, {and} \bibinfo{person}{Carsten Eickhoff}.} \bibinfo{year}{2021}\natexlab{}.
\newblock \showarticletitle{A transformer-based framework for multivariate time series representation learning}. In \bibinfo{booktitle}{\emph{Proceedings of the 27th ACM SIGKDD Conference on Knowledge Discovery \& Data Mining}}. \bibinfo{pages}{2114--2124}.
\newblock


\bibitem[Zhang et~al\mbox{.}(2020)]%
        {zhang2020self}
\bibfield{author}{\bibinfo{person}{Qiang Zhang}, \bibinfo{person}{Aldo Lipani}, \bibinfo{person}{Omer Kirnap}, {and} \bibinfo{person}{Emine Yilmaz}.} \bibinfo{year}{2020}\natexlab{}.
\newblock \showarticletitle{Self-attentive {H}awkes Process}. In \bibinfo{booktitle}{\emph{International Conference on Machine Learning}}. PMLR, \bibinfo{pages}{11183--11193}.
\newblock


\bibitem[Zhang et~al\mbox{.}(2022)]%
        {zhang2022self}
\bibfield{author}{\bibinfo{person}{Xiang Zhang}, \bibinfo{person}{Ziyuan Zhao}, \bibinfo{person}{Theodoros Tsiligkaridis}, {and} \bibinfo{person}{Marinka Zitnik}.} \bibinfo{year}{2022}\natexlab{}.
\newblock \showarticletitle{Self-supervised contrastive pre-training for time series via time-frequency consistency}.
\newblock \bibinfo{journal}{\emph{Advances in Neural Information Processing Systems}}  \bibinfo{volume}{35} (\bibinfo{year}{2022}), \bibinfo{pages}{3988--4003}.
\newblock


\bibitem[Zhuang et~al\mbox{.}(2021)]%
        {9134370}
\bibfield{author}{\bibinfo{person}{Fuzhen Zhuang}, \bibinfo{person}{Zhiyuan Qi}, \bibinfo{person}{Keyu Duan}, \bibinfo{person}{Dongbo Xi}, \bibinfo{person}{Yongchun Zhu}, \bibinfo{person}{Hengshu Zhu}, \bibinfo{person}{Hui Xiong}, {and} \bibinfo{person}{Qing He}.} \bibinfo{year}{2021}\natexlab{}.
\newblock \showarticletitle{A Comprehensive Survey on Transfer Learning}.
\newblock \bibinfo{journal}{\emph{Proc. IEEE}} \bibinfo{volume}{109}, \bibinfo{number}{1} (\bibinfo{year}{2021}), \bibinfo{pages}{43--76}.
\newblock
\urldef\tempurl%
\url{https://doi.org/10.1109/JPROC.2020.3004555}
\showDOI{\tempurl}


\bibitem[Zuo et~al\mbox{.}(2020)]%
        {zuo2020transformer}
\bibfield{author}{\bibinfo{person}{Simiao Zuo}, \bibinfo{person}{Haoming Jiang}, \bibinfo{person}{Zichong Li}, \bibinfo{person}{Tuo Zhao}, {and} \bibinfo{person}{Hongyuan Zha}.} \bibinfo{year}{2020}\natexlab{}.
\newblock \showarticletitle{Transformer {H}awkes process}. In \bibinfo{booktitle}{\emph{International Conference on Machine Learning}}. PMLR, \bibinfo{pages}{11692--11702}.
\newblock


\end{thebibliography}
